\documentclass{article}

\usepackage{hyperref}
\usepackage{url}
\usepackage{fullpage}

\usepackage{times}
\usepackage{amsmath,amssymb,amsthm}
\usepackage{multirow,graphicx,url}
\usepackage{algorithm,algorithmic,hyperref}
\usepackage{xspace}
\usepackage{stmaryrd}
\usepackage{array}
\usepackage{paralist}
\usepackage{enumitem}
\usepackage{booktabs}
\usepackage{subfig}
\usepackage{dsfont}
\usepackage{tabularx,epstopdf}
\usepackage{ textcomp }
\usepackage{caption}

\newtheorem{lem}{Lemma}
\newtheorem{thm}[lem]{Theorem}

\newcommand{\alg}{X1\xspace}

\newcommand{\norms}[1]{\|{#1}\|}
\newcommand{\reg}{\mu}

\newcommand{\lbar}{\bar L}
\renewcommand{\th}{\text{th}}

\makeatletter
\newcommand{\newreptheorem}[2]{\newtheorem*{rep@#1}{\rep@title} 
\newenvironment{rep#1}[1]{\def\rep@title{#2 \ref*{##1}}\begin{rep@#1}}{\end{rep@#1}}
}
\makeatother

\newreptheorem{lem}{Lemma}
\newreptheorem{thm}{Theorem}
\newreptheorem{clm}{Claim}

\newcommand{\R}{{\mathbb R}}
\newcommand{\Z}{{\mathbb Z}}
\newcommand{\RR}{{\cal R}}

\renewcommand{\H}{{\cal H}}
\newcommand{\X}{{\cal X}}
\newcommand{\D}{{\cal D}}
\renewcommand{\L}{{\cal L}}

\newcommand{\Y}{{\cal Y}}
\newcommand{\A}{{\cal A}}

\renewcommand{\P}{{\cal P}}

\newcommand{\ZZ}{{\cal Z}}

\newcommand{\bc}[1]{\left\{{#1}\right\}}
\newcommand{\br}[1]{\left({#1}\right)}
\newcommand{\bs}[1]{\left[{#1}\right]}
\newcommand{\abs}[1]{\left| {#1} \right|}
\newcommand{\norm}[1]{\left\| {#1} \right\|}

\newcommand{\bsd}[1]{\left\llbracket{#1}\right\rrbracket}

\renewcommand{\O}[1]{{\cal O}\br{{#1}}}

\newcommand{\E}[1]{{\mathbb E}\bsd{{#1}}}

\newcommand{\EE}[2]{\underset{#1}{\mathbb E}\bsd{{#2}}}
\newcommand{\Enob}[2]{\underset{#1}{\mathbb E}{{#2}}}

\newcommand{\ip}[2]{\left\langle{#1},{#2}\right\rangle}

\renewcommand{\vec}[1]{{\mathbf{#1}}}

\newcommand{\vecx}{\vec{x}}

\newcommand{\x}{\vec{x}}
\newcommand{\y}{\vec{y}}
\newcommand{\z}{\vec{z}}

\renewcommand{\v}{\vec{v}}

\newcommand{\tz}{\tilde \z}
\newcommand{\Lh}{\widehat{L}}

\newcommand{\Ind}[1]{{\mathbb I}\bs{{#1}}}

\def\vec#1{\mathchoice%
	{\mbox{\boldmath $\displaystyle\bf#1$}}
	{\mbox{\boldmath $\textstyle\bf#1$}}
	{\mbox{\boldmath $\scriptstyle\bf#1$}}
	{\mbox{\boldmath $\scriptscriptstyle\bf#1$}}}
\def\v#1{\protect\vec #1}

\title{Locally Non-linear Embeddings for Extreme Multi-label Learning}
\author{Kush Bhatia$^{*}$ \and Himanshu Jain$^{\#}$ \and Purushottam Kar$^{*}$ \and Prateek Jain$^{*}$ \and Manik Varma$^{*}$\\ $^{*}$Microsoft Research, Bangalore, INDIA\\\texttt{\{t-kushb, t-purkar, prajain, manik\}@microsoft.com}\\$^{\#}$Indian Institute of Technology, Delhi, INDIA\\
\texttt{himanshu.j689@gmail.com}}

\begin{document}

\maketitle

\begin{abstract}

The objective in extreme multi-label learning is to train a classifier that can automatically tag a novel data point with the most relevant {\it subset} of labels from an extremely large label set. Embedding based approaches make training and prediction tractable by assuming that the training label matrix is low-rank and hence the effective number of labels can be reduced by projecting the high dimensional label vectors onto a low dimensional linear subspace. Still, leading embedding approaches have been unable to deliver high prediction accuracies or scale to large problems as the low rank assumption is violated in most real world applications.

This paper develops the \alg classifier to address both limitations. The main technical contribution in \alg is a formulation for learning a small ensemble of local distance preserving embeddings which can accurately predict infrequently occurring (tail) labels. This allows \alg to break free of the traditional low-rank assumption and boost classification accuracy by learning embeddings which preserve pairwise distances between only the nearest label vectors.

We conducted extensive experiments on several real-world as well as benchmark data sets and compared our method against state-of-the-art methods for extreme multi-label classification. Experiments reveal that \alg can make significantly more accurate predictions then the state-of-the-art methods including both embeddings (by as much as 35\%) as well as trees (by as much as 6\%). \alg can also  scale efficiently to data sets with a million labels which are  beyond the pale of leading embedding methods.

\end{abstract}


\section{Introduction}
\label{sec:intro}

Our objective in this paper is to develop an extreme multi-label classifier, referred to as \alg, which can make significantly more accurate and faster predictions, as well as scale to larger problems, as compared to state-of-the-art embedding based approaches.

Extreme multi-label classification  addresses the problem of learning a classifier that can automatically tag a data point with the most relevant {\it subset} of labels from a large label set. For instance, there are more than a million labels (categories) on Wikipedia and one might wish to build a classifier that annotates a new article or web page with the subset of most relevant Wikipedia labels. It should be emphasized that multi-label learning is distinct from multi-class classification which aims to predict a single mutually exclusive label. 

Extreme multi-label  learning is a challenging research problem as one needs to simultaneously deal with hundreds of thousands, or even millions, of labels, features and training points. An obvious baseline is provided by the 1-vs-All technique where an independent classifier is learnt per label. Regrettably, this technique is infeasible due to the prohibitive training and prediction costs.  These problems could be ameliorated if a label hierarchy was provided. Unfortunately, such a hierarchy is unavailable in many applications~\cite{Agrawal13,Weston13}.

Embedding based approaches make training and prediction tractable by reducing the effective number of labels. Given a set of $n$ training points $\{(\v x_i,\v y_i)_{i=1}^{n}\}$ with $d$-dimensional feature vectors $\v x_i \in \R^d$ and $L$-dimensional label vectors $\v y_i \in \{0,1\}^L$, state-of-the-art embedding approaches  project the label vectors onto a lower $\widehat L$-dimensional linear subspace as $\z_i=\v U\v y_i$, based on a low-rank assumption. Regressors are then trained to predict $\z_i$ as $\v V \v x_i$. Labels for a novel point $\v x$ are predicted by post-processing $\v y = \v U^\dag \v V \v x$ where $\v U^\dag$ is a decompression matrix which lifts the embedded label vectors back to the original label space. 

Embedding methods mainly differ in the choice of compression and decompression techniques such as compressed sensing~\cite{Hsu09}, Bloom filters~\cite{Cisse13}, SVD~\cite{Tai10}, landmark labels~\cite{Balasubramanian12, Bi13}, output codes~\cite{Zhang11}, {\it etc}.  The state-of-the-art LEML algorithm~\cite{Yu14} directly optimizes for $\v U^\dag$, $\v V$  using the following objective: $\mbox{argmin}_{\v U^\dag, \v V}  Tr(\v U^{\dag\top} \v U^\dag) +  Tr(\v V^\top \v V) + 2C\sum_{i=1}^{n} \| y_i - \v U^\dag \v V \v x_i \|^2$. 

Embedding approaches have many advantages including simplicity, ease of implementation, strong theoretical foundations, the ability to handle label correlations, the ability to adapt to online and incremental scenarios, {\it etc}. Consequently, embeddings have proved to be the most popular approach for tackling extreme multi-label problems~\cite{Balasubramanian12, Bi13, Chen12, Cisse13, Feng11, Hsu09, Ji08, Yu14, Tai10, Weston11, Zhang11, Lin14}. 

Embedding approaches also have some limitations. They are slow at training and prediction even for a small embedding dimension $\widehat L$. For instance, on WikiLSHTC~\cite{Prabhu14,wikipedia-dataset}, a Wikipedia based challenge data set, LEML with $\widehat L=500$ took 22 hours for training even with  early termination while prediction took nearly 300 milliseconds per test point. In fact, for WikiLSHTC and other text applications with $\widehat{d}$-sparse feature vectors, LEML's prediction time $\Omega(\widehat L(\widehat{d} + L))$ can be an order of magnitude more than even 1-vs-All's prediction time $O(\widehat{d}L)$ (as $\widehat{d} = 42 \ll \widehat{L}=500$ for WikiLSHTC). 

\begin{figure}[t!]
\centering\hspace*{-10pt}
\begin{tabular}[h]{ccc}
\includegraphics[width=.33\textwidth]{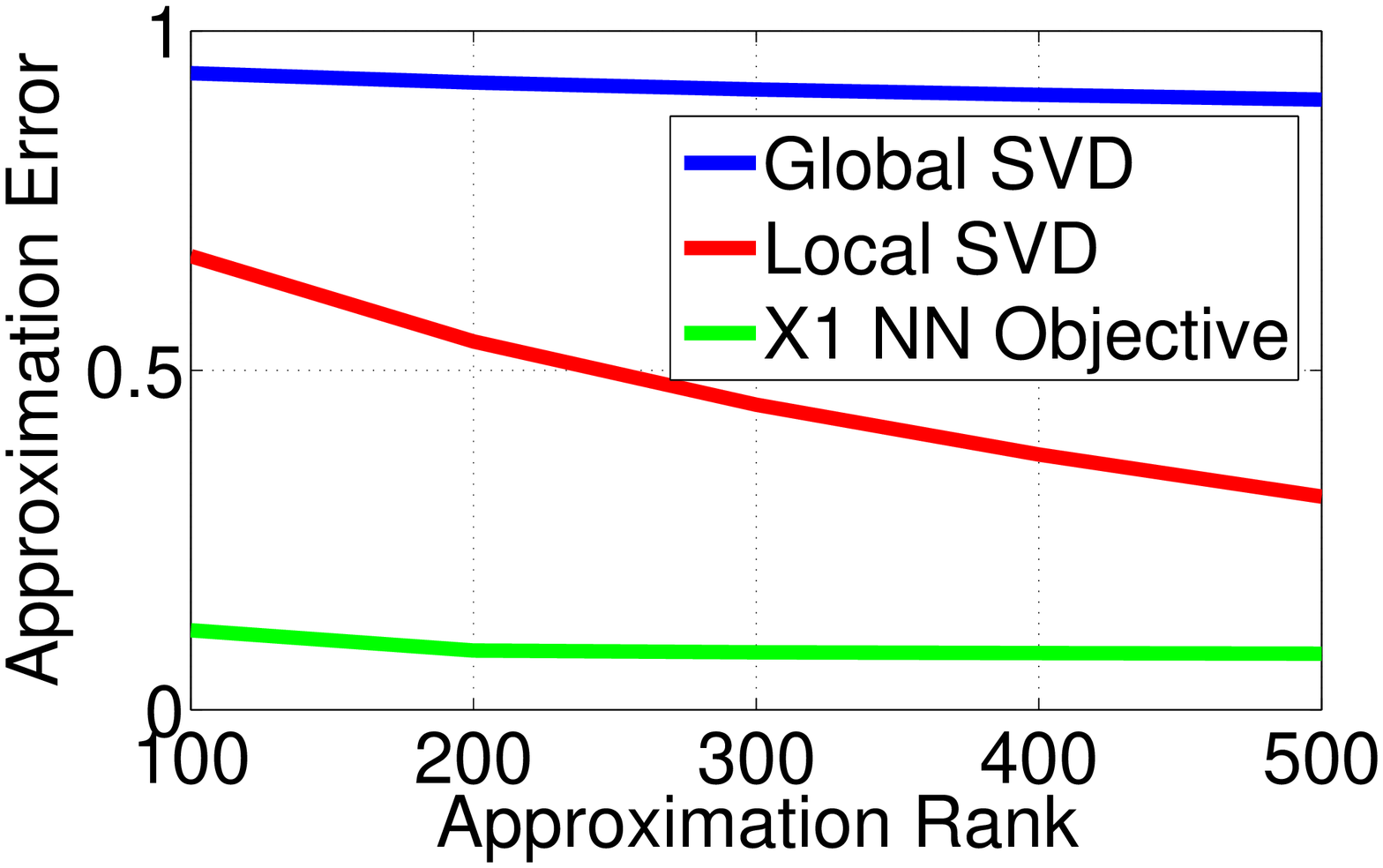}&
\includegraphics[width=.33\textwidth]{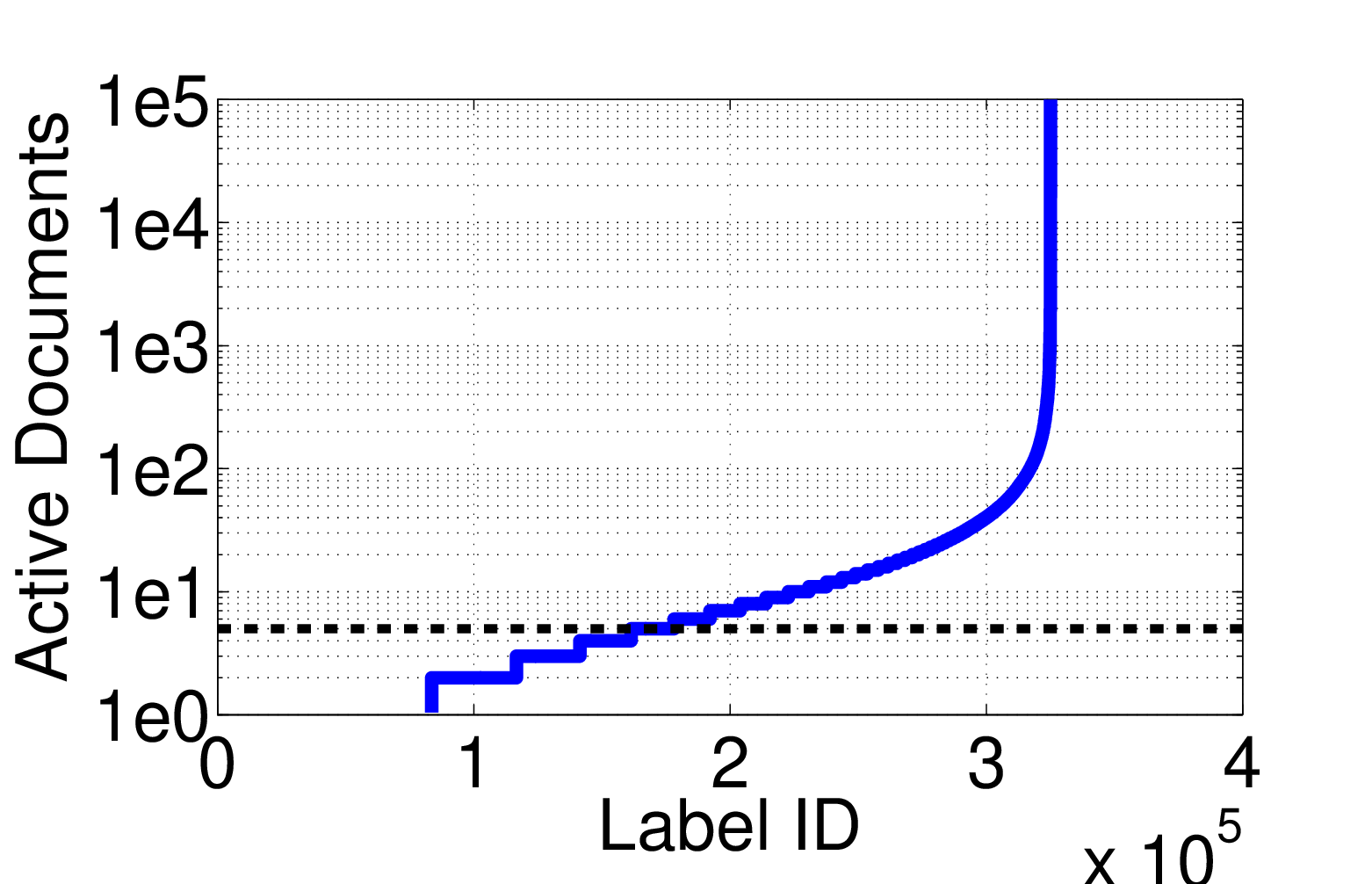}&
\includegraphics[width=.33\textwidth]{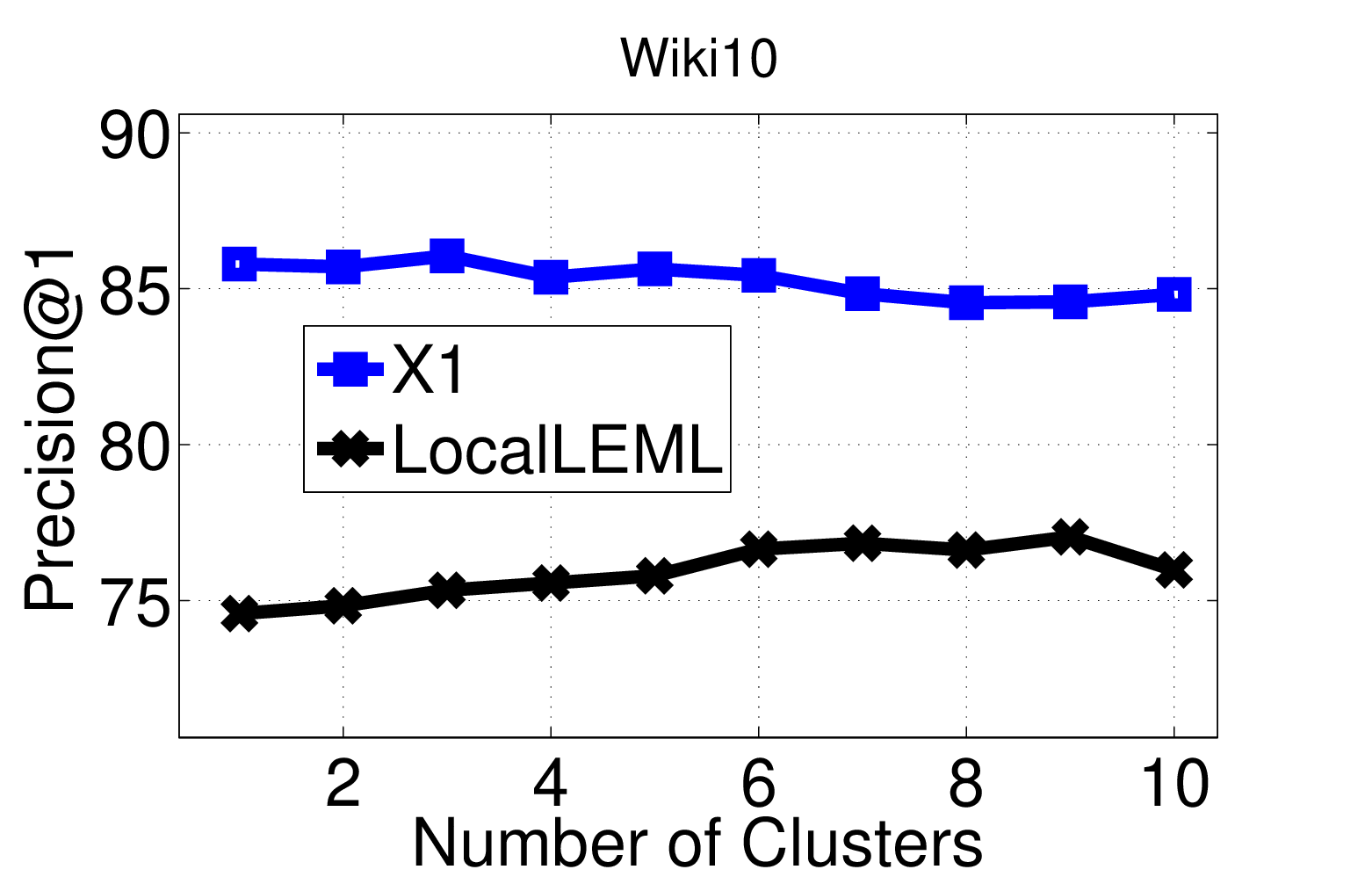}\\\vspace*{-2pt}
(a)&(b)&(c)\vspace*{-10pt}
\end{tabular}

\captionsetup{font=small}
\caption{(a) error $\|Y-Y_{\widehat L}\|_F^2/\|Y\|_F^2$ in approximating the label matrix $Y$. Global SVD denotes the error incurred by computing the rank ${\widehat L}$ SVD of $Y$. Local SVD computes rank ${\widehat L}$ SVD of $Y$ within each cluster. X1 NN objective denotes \alg's objective function. Global SVD incurs $90\%$ error and the error is decreasing at most linearly as well. (b) shows the number of documents in which each label is present for the WikiLSHTC data set. There are about $300K$ labels which are present in $<5$ documents lending it a `heavy tailed' distribution. (c) shows Precision@1 accuracy of \alg and localLEML on the Wiki-10 data set as we vary the number of clusters.}\vspace*{-10pt}
\label{fig:wikiERR}%
\end{figure}

More importantly, the critical assumption made by most embedding methods that the training label matrix is low-rank is violated in almost all real world applications.  Figure~\ref{fig:wikiERR}(a) plots the approximation error in the label matrix as $\widehat L$ is varied from $100$ to $500$ on the WikiLSHTC data set. As can be seen, even with a $500$-dimensional subspace the label matrix still has $90\%$ approximation error. We observe that this limitation arises primarily due to the presence of hundreds of thousands of ``tail" labels (see Figure~\ref{fig:wikiERR}(b)) which occur in at most $5$ data points each and, hence, cannot be well approximated by any linear low dimensional basis.

This paper develops the \alg algorithm which extends embedding methods in multiple ways to address these limitations. First, instead of projecting onto a linear low-rank subspace, \alg learns embeddings which non-linearly capture label correlations by preserving the pairwise distances between only the closest (rather than all) label vectors, i.\ e.\ $d(\z_i,\z_j) \approx d(\v y_i,\v y_j)$ if $i \in \textup{\textrm{kNN}}(j)$where $d$ is a distance metric. Regressors $\v V$ are trained to predict $\z_i =\v V \v x_i$. During prediction, rather than using a decompression matrix, \alg uses a k-nearest neighbour (kNN) classifier in the learnt embedding space, thus leveraging the fact that nearest neighbour distances have been preserved during training. Thus, for a novel point $\v x$, the predicted label vector is obtained as $\v y = \sum_{i: \v V \v x_i \in \textup{\textrm{kNN}}(\v V \v x)} \v y_i$. Our use of the kNN classifier is also motivated by the observation that kNN outperforms discriminative methods in acutely low training data regimes \cite{NgJ02} as in the case of tail labels.  

The superiority of \alg's proposed embeddings over traditional low-rank embeddings can be determined in two ways. First, as can be seen in Figure~\ref{fig:wikiERR}, the relative approximation error in learning \alg's embeddings is significantly smaller as compared to the low-rank approximation error. Second, \alg can improve over state-of-the-art embedding methods' prediction accuracy by as much as 35\% (absolute) on the challenging WikiLSHTC data set. \alg also significantly outperforms methods such as WSABIE~\cite{Weston11} which also use kNN classification in the embedding space but learn their embeddings using the traditional low-rank assumption.

However, kNN classifiers are known to be slow at prediction. \alg therefore clusters the training data into $C$ clusters, learns a separate embedding per cluster and performs kNN classification within the test point's cluster alone. This reduces \alg's prediction costs to $O(\widehat{d}C + \widehat{d}\widehat{L} + N_C\widehat{L})$ for determining the cluster membership of the test point, embedding it and then performing kNN classification respectively, where $N_C$ is the number of points in the cluster to which the test point was assigned. \alg can therefore be more than two orders of magnitude faster at prediction than LEML and other embedding methods on the WikiLSHTC data set where $C = 300, N_C \lesssim 13K, \widehat{L}_{\mbox{\alg}} = 50$ and $\widehat{D} = 42$. Clustering can also reduce \alg's training time by almost a factor of $C$. This allows \alg to scale to the Ads1M data set involving a million labels which is beyond the pale of leading embedding methods. 

Of course, clustering is not a significant technical innovation in itself, and could easily have been applied to traditional embedding approaches. However, as our results demonstrate, state-of-the-art methods such as LEML do not benefit much from clustering. Clustered LEML's prediction accuracy continues to lag behind \alg's by $14\%$ on WikiLSHTC and the training time on Ads1M continues to be prohibitively large.

The main limitation of clustering is that it can be unstable in high dimensions. \alg compensates by learning a small ensemble where each individual learner is generated by a different random clustering. This was empirically found to help tackle instabilities of clustering and significantly boost prediction accuracy with only linear increases in training and prediction time. For instance, on WikiLSHTC, \alg's prediction accuracy was $56\%$ with an 8 millisecond prediction time whereas LEML could only manage $20\%$ accuracy while taking 300 milliseconds for prediction per test point.

Recently, tree based methods~\cite{Agrawal13, Prabhu14, Weston13} have also become popular for extreme multi-label learning as they enjoy significant accuracy gains over the existing embedding methods. For instance, FastXML~\cite{Prabhu14} can achieve a prediction accuracy of $49\%$ on WikiLSHTC using a 50 tree ensemble. However, \alg is now able to extend embedding methods to outperform tree ensembles, achieving 49.8\% with 2 learners and 55\% with 10. Thus, by learning local distance preserving embeddings, \alg can now obtain the best of both worlds. In particular, \alg can achieve the highest prediction accuracies across all methods on even the most challenging data sets while retaining all the benefits of embeddings and eschewing the disadvantages of large tree ensembles such as large model size and lack of theoretical understanding.

Our contributions in this paper are: First, we identify that the low-rank assumption made by most embedding methods is violated in the real world and that local distance preserving embeddings can offer a superior alternative. Second, we propose a novel formulation for learning such embeddings and show that it has sound theoretical properties. In particular, we prove that \alg consistently preserves nearest neighbours in the label space and hence learns good quality embeddings. Third, we build an efficient pipeline for training and prediction which can be orders of magnitude faster than state-of-the-art embedding methods while being significantly more accurate as well.

\vspace*{-5pt}
\section{Method}\vspace*{-5pt}
\label{sec:method}

Let ${\cal D}=\{(\x_1, \y_1) \dots (\x_n, \y_n)\}$ be the given training data set, $\x_i \in {\cal X}\subseteq \R^d$ be the input feature vector, $\y_i\in {\cal Y}\subseteq \{0,1\}^L$ be the corresponding label vector, and $y_{ij}=1$ iff the $j$-th label is turned on for $\x_i$. Let $X=[\x_1, \dots, \x_n]$ be the data matrix and $Y=[\y_1,\dots,\y_n]$ be the label matrix. Given ${\cal D}$, the goal is to learn a  multi-label classifier $f:\R^d\rightarrow \{0,1\}^L$ that accurately predicts the label vector for a given test point. Recall that in extreme multi-label settings, $L$ is very large and is of the same order as $n$ and $d$, ruling out several standard approaches such as 1-vs-All.



We now present our algorithm \alg which is designed primarily to scale efficiently for large $L$. Our algorithm is an embedding-style algorithm, i.e., during training we  map the label vectors $\y_i$ to $\widehat L$-dimensional vectors $\z_i\in \R^{\widehat L}$ and  learn a set of regressors $V\in \R^{\widehat{L}\times d}$ s.t. $\z_i \approx V \x_i, \forall i$. During the test phase, for an unseen point $\x$, we first compute its embedding $V\x$ and then perform kNN over the set $[Vx_1, Vx_2, \dots, Vx_n]$. To scale our algorithm, we perform a clustering of all the training points and apply the above mentioned procedures in each of the cluster separately. Below, we first discuss our method to compute the embeddings $\z_i$s and the regressors $V$. Section~\ref{sec:scale} then discusses our approach for scaling the method to large data sets. 

\begin{figure}[t!]
\begin{minipage}[t]{0.495\linewidth}
\begin{algorithm}[H]
   \caption{\small \alg : Train Algorithm}
   \label{alg:train1}
\begin{algorithmic}[1]
	\small{
   \REQUIRE ${\cal D}=\{(x_1, y_1) \dots (x_n, y_n)\}$,  embedding dimensionality: $\Lh$, no. of neighbors: $\bar{n}$, no. of clusters: $C$, regularization parameter: $\lambda, \reg$, L1 smoothing parameter $\rho$
   \STATE Partition $X$ into $Q^1, .., Q^C$ using $k$-means 
   \FOR {each partition $Q^j$}
   		\STATE Form $\Omega$ using $\bar{n}$ nearest neighbors of each label vector $\y_i \in Q^j$
   		\STATE $\left[ U\ \Sigma\right] \leftarrow \text{SVP}(P_{\Omega}(Y^{j}{Y^{j}}^T),\ \Lh)$ 
   		\STATE $Z^j \leftarrow U\Sigma^{\frac{1}{2}}$ 
   		\STATE $V^j\leftarrow ADMM(X^j, Z^j, \lambda,\ \reg,\ \rho)$
                \STATE $Z^j=V^j X^j$
   \ENDFOR
   \STATE {\bf Output}: $\{(Q^1, V^1, Z^1),\dots, (Q^C, V^C, Z^C\}$
	}
\end{algorithmic}
\end{algorithm}
\vspace{-3.5ex}
\floatname{algorithm}{Algorithm}
\begin{algorithm}[H]
   \caption{\small \alg : Test Algorithm}
   \label{alg:test1}
\begin{algorithmic}[1]
	\small{
   	\REQUIRE Test point: $\x$, no. of NN: $\bar{n}$, no. of desired labels: $p$
   	\STATE $Q_{\tau}$: partition closest to $\x$
   	\STATE $\z \leftarrow V^{\tau}\x$
        \STATE ${\cal N}_z\leftarrow$ $\bar{n}$ nearest neighbors of $z$ in $Z^\tau$
        \STATE $P_x\leftarrow$ empirical label dist. for points $\in{\cal N}_z$
  	\STATE $y_{pred} \leftarrow Top_{p}(P_x)$
	}
\end{algorithmic}
\end{algorithm}
\end{minipage}
\begin{minipage}[t]{0.495\linewidth}
\floatname{algorithm}{\phantom{g}Sub-routine}
\begin{algorithm}[H]
   \caption{\small \alg : SVP}
   \label{alg:train2}
\begin{algorithmic}[1]
	\small{
  	\REQUIRE Observations: $G$, index set: $\Omega$, dimensionality: $\Lh$
	\STATE $M_1 := 0$, $\eta=1$
	\REPEAT
        \STATE $\widehat{M}\leftarrow M+\eta (G-P_{\Omega}(M))$
		\STATE $ [U\ \Sigma]\leftarrow\text{Top-EigenDecomp}(\widehat{M},\ \Lh)$
                \STATE $\Sigma_{ii}\leftarrow\max(0, \Sigma_{ii}), \forall i$
		\STATE $M \leftarrow U\cdot \Sigma \cdot U^T$
	\UNTIL{Convergence}
	\STATE {\bf Output}: $U$, $\Sigma$
	}
\end{algorithmic}
\end{algorithm}
\vspace{-3.6ex}
\floatname{algorithm}{\phantom{g}Sub-routine}
\begin{algorithm}[H]
   \caption{\small \alg : ADMM }
   \label{alg:ADMM}
\begin{algorithmic}[1]
	\small{
  	\REQUIRE Data Matrix : $X$, Embeddings : $Z$, Regularization Parameter : $\lambda, \reg$, Smoothing Parameter : $\rho$
	\STATE $\beta := 0, \alpha := 0$
	\REPEAT
		\STATE $Q \leftarrow (Z + \rho (\alpha - \beta))X^\top$
        \STATE $V \leftarrow Q(XX^\top(1+\rho)+\lambda I)^{-1} $
        \STATE $\alpha \leftarrow(VX + \beta)$
        \STATE $\alpha_i = \mbox{sign}(\alpha_i)\cdot \max(0, |\alpha_i|-\frac{\reg}{\rho})$, $\forall i$
        \STATE $\beta \leftarrow \beta +VX - alpha$
	\UNTIL{Convergence}
	\STATE {\bf Output}: $V$
	}
\end{algorithmic}
\end{algorithm}

\end{minipage}
\end{figure}
\subsection{Learning Embeddings}\vspace*{-5pt}
As mentioned earlier, our approach is motivated by the fact that a typical real-world data set tends to have a large number of tail labels that ensure that the label matrix $Y$ cannot be well-approximated using a low-dimensional linear subspace (see Figure~\ref{fig:wikiERR}). However, $Y$ can still be accurately modeled using a low-dimensional non-linear manifold. That is, instead of preserving distances (or inner products) of a given label vector to all the training points, we attempt to preserve the distance to only a few nearest neighbors. That is, we wish to find a $\Lh$-dimensional embedding matrix $Z=[\z_1, \dots, \z_n]\in \R^{\Lh\times n}$ which minimizes the following objective: 
\begin{align}
  \min_{Z\in \R^{\Lh\times n}}\|P_{\Omega}(Y^TY)-P_{\Omega}(Z^TZ)\|_F^2+\lambda \|Z\|_1,\label{eq:nl1}
\end{align}
where the index set $\Omega$ denotes the set of neighbors that we wish to preserve, i.e., $(i,j)\in \Omega$ iff $j\in {\cal N}_i$. ${\cal N}_i$ denotes a set of nearest neighbors of $i$. We select ${\cal N}_i=\arg\max_{S, |S|\leq \alpha\cdot n}\sum_{j\in S}(\y_i^T\y_j)$, which is the set of $\alpha\cdot n$ points with the largest inner products with $\y_i$. 
$P_{\Omega}:\R^{n\times n}\rightarrow \R^{n\times n}$ is defined as:
\begin{equation}
  (P_{\Omega}(Y^TY))_{ij}=\begin{cases}\ip{\y_i}{\y_j},& \text{ if } (i,j)\in \Omega,\\
0,& \text {otherwise}.\end{cases}
\end{equation}
Also, we add $L_1$ regularization, $\|Z\|_1=\sum_{i}\|\z_i\|_1$, to the objective function to obtain sparse embeddings. Sparse embeddings have three key advantages: a) they reduce prediction time, b) reduce the size of the model, and c) avoid overfitting.  Now, given the embeddings $Z=[\z_1, \dots, \z_n]\in \R^{\Lh\times n}$, we wish to learn a multi-regression model to predict the embeddings $Z$ using the input features. That is, we require that $Z\approx VX$ where $V\in \R^{\Lh\times d}$. Combining the two formulations and adding an $L_2$-regularization for $V$, we get: 
\begin{align}
  \min_{V\in \R^{\Lh\times d}}\|P_{\Omega}(Y^TY)-P_{\Omega}(X^TV^TVX)\|_F^2+\lambda \|V\|_F^2+\mu \|VX\|_1.\label{eq:nl2}
\end{align}
Note that the above problem formulation is somewhat similar to a few existing methods for non-linear dimensionality reduction that also seek to preserve distances to a few near neighbors \cite{MVU,MVE}. However, in contrast to our approach, these methods do not have a direct out of sample generalization, do not scale well to large-scale data sets, and lack rigorous generalization error bounds. 

{\bf Optimization}: 
We first note that optimizing \eqref{eq:nl2} is a significant challenge as the objective function is non-convex as well as non-differentiable. Furthermore, our goal is to perform optimization for data sets where $L, n, d \gg 100,000$. To this end, we divide the optimization into two phases. We first learn embeddings $Z=[\z_1,\dots, \z_n]$ and then learn regressors $V$ in the second stage. That is, $Z$ is obtained by  directly solving \eqref{eq:nl1} but without the $L_1$ penalty term: \vspace*{-12pt}

\begin{align}
\min_{Z, Z\in \R^{\Lh\times n}} \|P_{\Omega}(Y^TY)-P_{\Omega}(Z^TZ)\|_F^2\equiv \min_{\substack{M\succeq 0,\\ rank(M)\leq \Lh}}\|P_{\Omega}(Y^TY)-P_{\Omega}(M)\|_F^2,\label{eq:nl4}
\end{align}
where $M=Z^TZ$. Next, $V$ is obtained by solving the following problem:
\begin{align}
  \min_{V\in \R^{\Lh\times d}}\|Z-VX\|_F^2+\lambda \|V\|_F^2 + \mu \|VX\|_1.\label{eq:nl5}
\end{align}
Note that the $Z$ matrix obtained using \eqref{eq:nl4} need not be sparse. However, we store and use $VX$ as our embeddings, so that sparsity is still maintained.

{\em Optimizing \eqref{eq:nl4}:} Note that even the simplified problem \eqref{eq:nl4} is an instance of the popular low-rank matrix completion problem and is known to be NP-hard in general. The main challenge arises due to the non-convex  rank constraint on $M$. However, using the Singular Value Projection (SVP) method \cite{JainMD10}, a popular matrix completion method, we can guarantee convergence to a local minima. 

SVP is a simple projected gradient descent method where the projection is onto the set of low-rank matrices. That is, the $t$-th step update for SVP is given by: 
\begin{equation}
  \label{eq:svp1}
  M_{t+1}=P_{\Lh}(M_t+\eta P_{\Omega}(Y^TY-M_t)),
\end{equation}
where $M_t$ is the $t$-th step iterate, $\eta>0$ is the step-size, and $P_{\Lh}(M)$ is the projection of $M$ onto the set of rank-$\Lh$ positive semi-definite definite (PSD) matrices. Note that while the set of rank-$\Lh$ PSD matrices is non-convex, we can still project onto this set efficiently using the eigenvalue decomposition of $M$. That is, if $M=U_M \Lambda_M U_M^T$ be the eigenvalue decomposition of $M$. Then, $$P_{\Lh}(M)=U_M(1:r) \cdot \Lambda_M(1:r)\cdot U_M(1:r)^T,$$ where $r=\min(\Lh, \Lh^+_M)$ and $\Lh^+_M$ is the number of positive eigenvalues of $M$. $\Lambda_M(1:r)$ denotes the top-$r$ eigenvalues of $M$ and $U_M(1:r)$ denotes the corresponding eigenvectors. 

While the above update restricts the rank of all intermediate iterates $M_t$ to be at most $\Lh$,  computing rank-$\Lh$ eigenvalue decomposition  can still be fairly expensive for large $n$. However, by using special structure in the update \eqref{eq:svp1}, one can significantly reduce eigenvalue decomposition's computation complexity as well. In general, the eigenvalue decomposition can be computed in time $O(\Lh \zeta)$ where $\zeta$ is the time complexity of computing a matrix-vector product. Now, for SVP update \eqref{eq:svp1}, matrix has special structure of $\hat{M}=M_t+\eta P_{\Omega}(Y^TY-M_t)$. Hence $\zeta=O(n\Lh+n\bar{n})$ where $\bar{n}=|\Omega|/n^2$ is the average number of neighbors preserved by \alg. Hence, the per-iteration time complexity reduces to $O(n\Lh^2+n\Lh\bar{n})$ which is linear in $n$, assuming $\bar{n}$ is nearly constant.

{\em Optimizing \eqref{eq:nl5}:} \eqref{eq:nl5} contains an $L_1$ term which makes the problem non-smooth. Moreover, as the $L_1$ term  involves both $V$ and $X$, we cannot directly apply the standard prox-function based algorithms. Instead, we use the ADMM method to optimize \eqref{eq:nl5}. See Sub-routine~\ref{alg:ADMM} for the updates and ~\cite{SprechmannLYBS13} for a detailed derivation of the algorithm.

{\bf Generalization Error Analysis}: 
Let $\P$ be a fixed (but unknown) distribution over ${\cal X}\times {\cal Y}$. Let each training point $(\x_i, \y_i)\in {\cal D}$ be sampled i.i.d. from $\P$. Then, the goal of our non-linear embedding method \eqref{eq:nl2} is to learn an embedding matrix $A=V^TV$ that preserves nearest neighbors (in terms of label distance/intersection) of any $(\x,\y)\sim \P$. 
The above requirements can be formulated as the following stochastic optimization problem: 
\begin{align}
\label{eq:nl_embed_stoch}
\min_{\substack{A\succeq 0\\rank(A)\leq k}} \L(A)=\Enob{(\x,\y), (\vec{\widetilde{x}},\vec{\widetilde{y}})\sim \P}{\ell(A;(\x,\y),(\vec{\widetilde{x}},\vec{\widetilde{y}}))},
\end{align}
where the loss function  $\ell(A;(\x,\y),(\vec{\widetilde{x}},\vec{\widetilde{y}})) = g(\ip{\vec{\widetilde{y}}}{\y})({\ip{\vec{\widetilde{y}}}{\y}-\vec{\widetilde{x}}^T A\x})^2 $, and $g(\ip{\vec{\widetilde{y}}}{\y}) = \Ind{\ip{\vec{\widetilde{y}}}{\y}\geq \tau}$, where $\Ind{\cdot}$ is the indicator function. Hence, a loss is incurred only if $\y$ and $\tilde{\y}$ have a large inner product. For an appropriate selection of the neighborhood selection operator $\Omega$,  \eqref{eq:nl2}  indeed minimizes a regularized empirical estimate of the loss function \eqref{eq:nl_embed_stoch}, i.e., it is a regularized ERM w.r.t. \eqref{eq:nl_embed_stoch}.


We now show that the optimal solution $\widehat{A}$ to \eqref{eq:nl2} indeed minimizes the loss \eqref{eq:nl_embed_stoch} upto an additive approximation error. The existing techniques for analyzing excess risk in stochastic optimization  require the empirical loss function to be decomposable over the training set, and as such do not apply to \eqref{eq:nl2} which contains loss-terms with two training points. Still, using techniques from the AUC maximization literature \cite{KarSJK13}, we can provide interesting excess risk bounds for Problem \eqref{eq:nl_embed_stoch}.

\begin{thm}
\label{thm:plain-excess-risk-bd}
With probability at least $1 - \delta$ over the sampling of the dataset $\D$, the solution $\hat A$ to the optimization problem \eqref{eq:nl2} satisfies\vspace*{-8pt}
$$
\L(\hat A) \leq \inf_{A^\ast\in\A}\Big\{\L(A^\ast) + \overbrace{C\br{\lbar^2 + \br{r^2+\norms{A^\ast}_F^2}R^4}\sqrt{\frac{1}{n}\log\frac{1}{\delta}}}^{\text{E-Risk}(n)}\Big\},\vspace*{-5pt}$$
where $\hat A$ is the minimizer of \eqref{eq:nl2}, $r=\frac{\lbar}{\lambda}$ and $\A := \bc{A \in \R^{d\times d}:A \succeq 0, rank(A) \leq \Lh}$.  
\end{thm}
See Appendix~\ref{app:genbound} for a proof of the result. Note that the generalization error bound is {\em independent} of both $d$ and $L$, which is critical for extreme multi-label classification problems with large $d, L$. In fact, the error bound is only dependent on $\lbar\ll L$, which is the average number of positive labels per data point. Moreover, our bound also provides a way to compute best regularization parameter $\lambda$ that minimizes the error bound. However, in practice, we set $\lambda$ to be a fixed constant. 

Theorem~\ref{thm:plain-excess-risk-bd} only preserves the {\em population} neighbors of a test point. Theorem~\ref{thm:excess-risk-to-train-bd}, given in Appendix~\ref{app:genbound}, extends Theorem~\ref{thm:plain-excess-risk-bd} to ensure that the neighbors in the {\em training} set are also preserved. 
We would also like to stress that our excess risk bound is universal and hence holds even if $\hat A$ does not minimize \eqref{eq:nl2}, i.e., $\L(\hat A)\leq \L(A^*)+\text{E-Risk}(n)+(\L(\hat A)-\L(\hat(A^*)),$ where $\text{E-Risk}(n)$ is given in Theorem~\ref{thm:plain-excess-risk-bd}. 


\vspace*{-5pt}
\subsection{Scaling to Large-scale Data sets}\vspace*{-5pt}
\label{sec:scale}
For large-scale data sets, one might require the embedding dimension $\Lh$ to be fairly large (say a few hundreds) which might make computing the updates \eqref{eq:svp1} infeasible. Hence, to scale to such large data sets, \alg clusters the given datapoints into smaller local region. Several text-based data sets indeed reveal that there exist small local regions in the feature-space where the number of points as well as the number of labels is reasonably small. Hence, we can train our embedding method over such local regions without significantly sacrificing overall accuracy. 

We would like to stress that despite clustering datapoints in homogeneous regions, the label matrix of any given cluster is still not close to low-rank. Hence, applying a state-of-the-art linear embedding method, such as LEML, to each cluster is still significantly less accurate when compared to our method (see Figure~\ref{fig:wikiERR}). Naturally, one can cluster the data set into an extremely large number of regions, so that eventually the label matrix is low-rank in each cluster. However, increasing the number of clusters beyond a certain limit might decrease accuracy as the error incurred during the cluster assignment phase itself might nullify the gain in accuracy due to better embeddings. Figure~\ref{fig:wikiERR} illustrates this phenomenon where increasing the number of clusters beyond a certain limit in fact decreases accuracy of LEML. 

Algorithm~\ref{alg:train1} provides a pseudo-code of our training algorithm. We first cluster the datapoints into $C$ partitions. Then, for each partition we learn a set of embeddings using Sub-routine~\ref{alg:train2} and then compute the regression parameters $V^\tau, 1\leq \tau\leq C$ using Sub-routine~\ref{alg:ADMM}. For a given test point $\x$, we first find out the appropriate cluster $\tau$. Then, we find the embedding $\z=V^\tau\x$. The label vector is then predicted using $k$-NN in the embedding space. See Algorithm~\ref{alg:test1} for more details.

Owing to the curse-of-dimensionality, clustering turns out to be quite unstable for data sets with large $d$ and in many cases leads to some drop in prediction accuracy. To safeguard against such instability, we use an ensemble of models generated using different sets of clusters. We use different initialization points in our clustering procedure to obtain different sets of clusters. Our empirical results demonstrate that using such ensembles leads to significant increase in accuracy of \alg (see Figure~\ref{fig:learnerP1}) and also leads to stable solutions with small variance (see Table~\ref{tab:wikiStab}).


\section{Experiments}
\label{sec:exp}
\begin{figure}[t!]
  \centering\hspace*{-10pt}
\begin{tabular}{ccc}
  \includegraphics[width=.33\textwidth]{Plots_new/Appendix/wikiLSHTC_p1}\hspace*{-3pt}&
  \includegraphics[width=.33\textwidth]{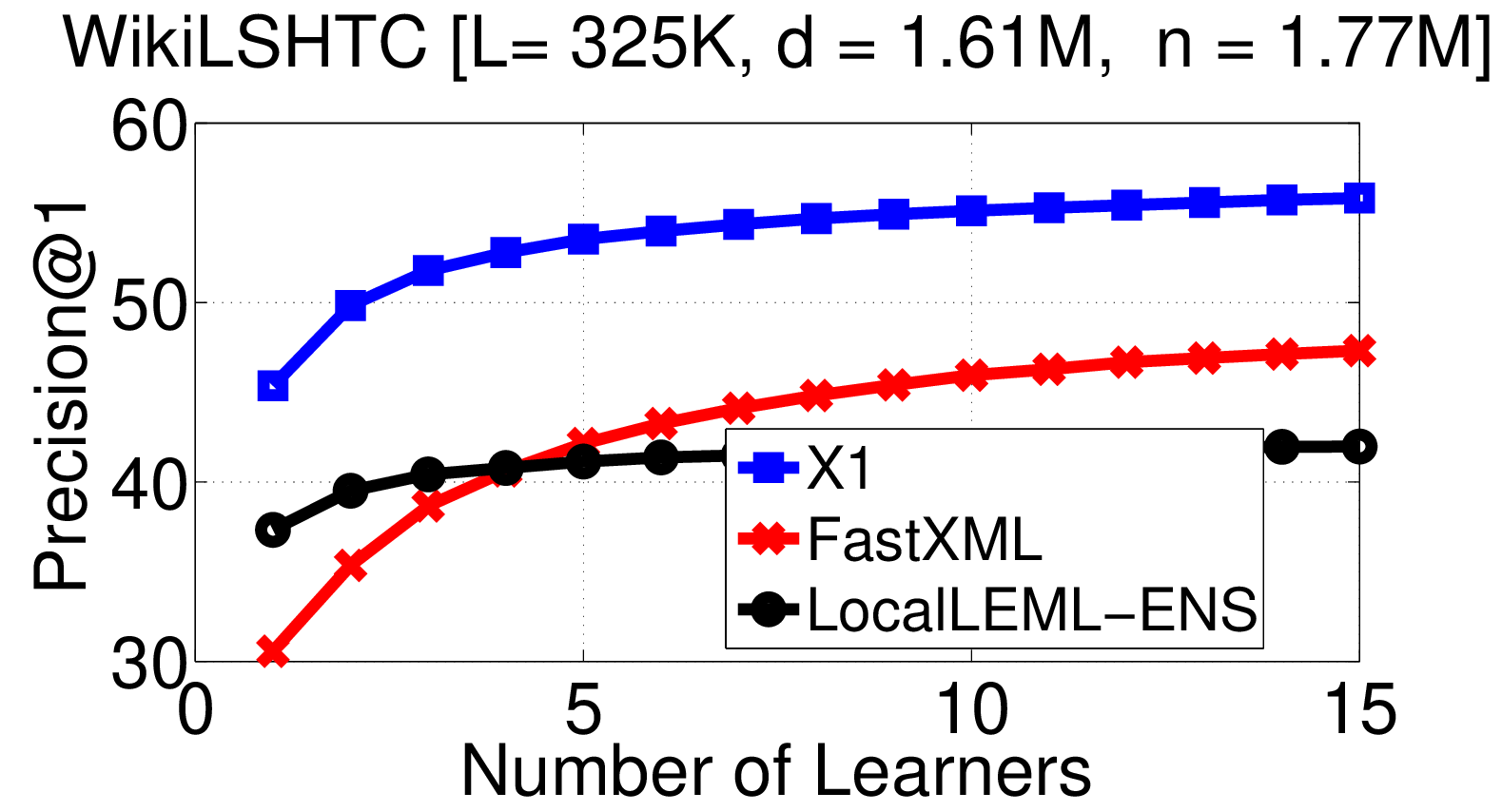}\hspace*{-2pt}&
  \includegraphics[width=.33\textwidth]{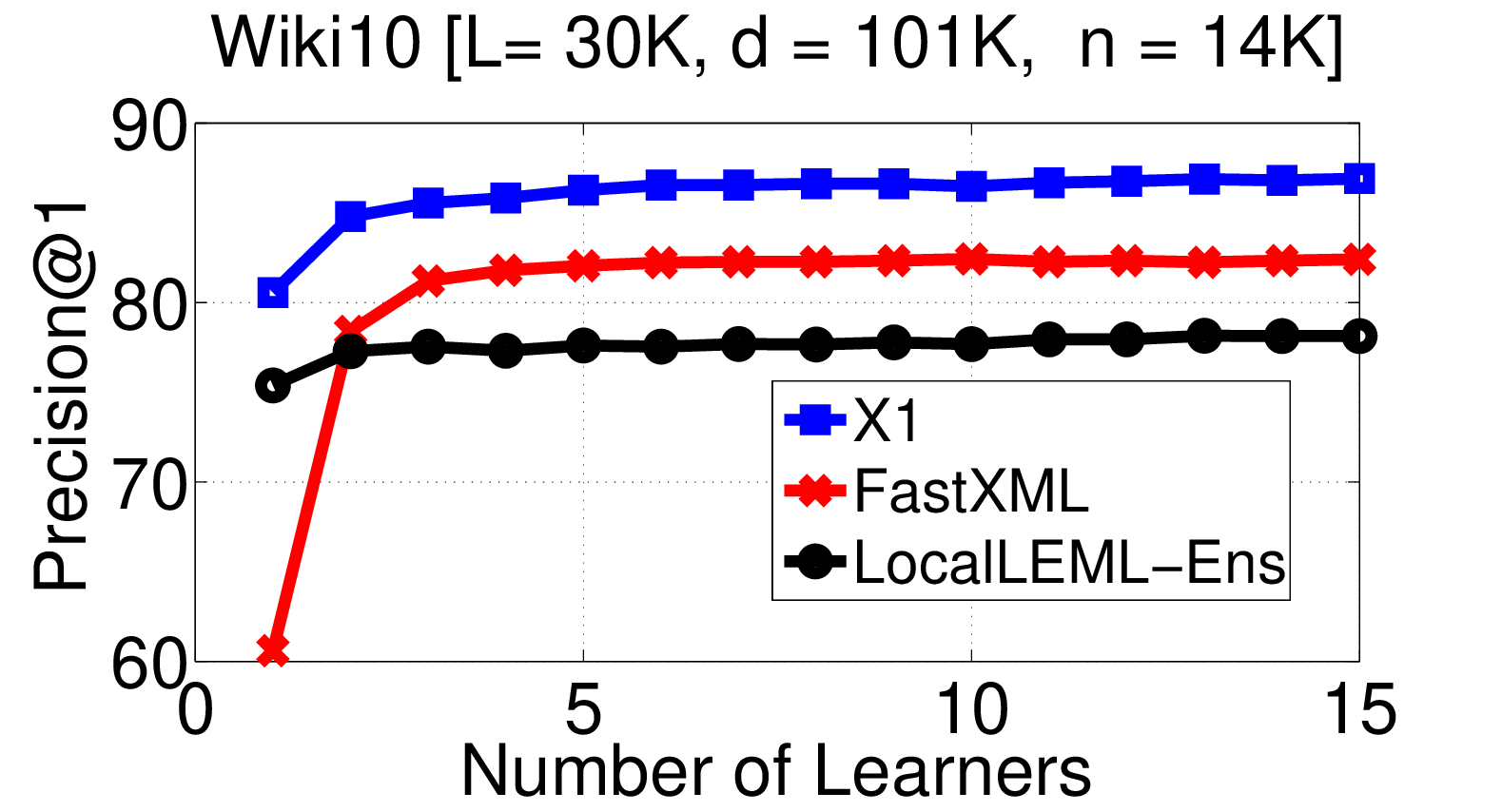}\\
(a)&(b)&(c)\vspace*{-10pt}
\end{tabular}
\captionsetup{font=small}
\caption{Variation in Precision@1 accuracy with model size and the number of learners on large-scale data sets. Clearly, \alg achieves better accuracy than FastXML and LocalLEML-Ensemble at every point of the curve. For WikiLSTHC, \alg with a single learner is more accurate than LocalLEML-Ensemble with even 15 learners. Similarly, \alg with $2$ learners achieves more accuracy than FastXML with $50$ learners.}
  \label{fig:learnerP1}\vspace*{-10pt}
\end{figure}
Experiments were carried out on some of the largest  extreme multi-label benchmark data sets demonstrating that \alg could achieve significantly higher prediction accuracies as compared to the state-of-the-art. It is also demonstrated that \alg could be faster at training and prediction than leading embedding techniques such as LEML.

\textbf{Data sets}: Experiments were carried out on multi-label data sets including Ads1M~\cite{Prabhu14} (1M labels), Amazon~\cite{Snapnets} (670K labels), WikiLSHTC (320K labels),  DeliciousLarge~\cite{Wetzker08} (200K labels) and Wiki10~\cite{Zubiaga09} (30K labels). All the data sets are publically available except  Ads1M which is proprietary and is included here to test the scaling capabilities of \alg.

Unfortunately, most of the existing embedding techniques do not scale to such large data sets. We therefore also present comparisons on publically available small data sets such as BibTeX~\cite{Katakis08}, MediaMill~\cite{Snoek06}, Delicious~\cite{Tsoumakas08} and EURLex~\cite{Mencia08}. Table \ref{tab:stats} in the supplementary material lists the statistics of each of these data sets. 

\textbf{Baseline algorithms}: This paper's primary focus is on comparing \alg to state-of-the-art methods which can scale to the large data sets such as embedding based LEML~\cite{Yu14} and tree based  FastXML~\cite{Prabhu14} and LPSR~\cite{Weston13}. Na\"ive Bayes was used as the base classifier in LPSR as was done in~\cite{Prabhu14}. Techniques such as  CS~\cite{Hsu09}, CPLST~\cite{ChenL12}, ML-CSSP~\cite{Bi13},  1-vs-All~\cite{Hariharan12} could only be trained on the small data sets given standard resources. Comparisons between \alg and such techniques are therefore presented in the supplementary material. The implementation for LEML and FastXML was provided by the authors. We implemented the remaining algorithms and ensured that the published results could be reproduced and were verified by the authors wherever possible.

\textbf{Hyper-parameters}: Most of \alg's hyper-parameters  were kept fixed including the number of clusters in a learner $\left( \lfloor{N_{\mbox{Train}}/6000}\rfloor \right)$, embedding dimension (100 for the small data sets and 50 for the large), number of learners in the ensemble (15), and the parameters used for optimizing \eqref{eq:nl2}. The remaining two hyper-parameters, the $k$ in kNN and the number of neighbours  considered during SVP, were both set by limited validation on a validation set.

The hyper-parameters for all the other algorithms were set using fine grained validation on each data set so as to achieve the highest possible prediction accuracy for each method. In addition, all the embedding methods were allowed a much larger embedding dimension ($0.8 L$) than \alg (100) to give them as much opportunity as possible to outperform \alg.

\textbf{Evaluation Metric}: Precision at $k$ (P@k)has been widely adopted as the metric of choice for evaluating extreme multi-label algorithms~\cite{Agrawal13, Hsu09, Prabhu14, Weston11, Weston13, Yu14}. This is motivated by real world application scenarios such as tagging and recommendation. Formally, the precision at $k$ for a predicted score vector $\hat{\v y} \in {\cal R}^L$ is the fraction of correct positive predictions in the top $k$ scores of $\hat{\v y}$.

\textbf{Results on large data sets with more than 100K labels}:. Table \ref{tab:expResLSa} compares \alg's prediction accuracy, in terms of P@k (k$=\{1,3,5\}$), to all the leading methods that could be trained on five such data sets. \alg could improve over the leading embedding method, LEML, by as much as 35\% and 15\% in terms of P@1 and P@5 on the  WikiLSHTC data set. Similarly, \alg  outperformed LEML by 
27\% and 22\% in terms of P@1 and P@5 on the Amazon data set which also has many tail labels. The gains on the other data sets are consistent, but smaller, as the tail label problem is not so acute. \alg could also outperform the leading tree method, FastXML, by 6\% in terms of both P@1 and P@5 on WikiLSHTC and Wiki10 respectively. This demonstrates the superiority of \alg's overall pipeline constructed using local distance preserving embeddings followed by kNN classification.

\alg also has better scaling properties as compared to all other embedding methods. In particular, apart from LEML, no other embedding approach could scale to the large data sets and, even LEML could not scale to Ads1M with a million labels. In contrast, a single \alg learner could be learnt on WikiLSHTC in 4 hours on a single core and already gave $\sim$ 20\% improvement in P@1 over LEML (see Figure~\ref{fig:learnerP1} for the variation in accuracy vs \alg learners).  In fact, \alg's training time on WikiLSHTC was comparable to that of tree based FastXML. FastXML trains 50 trees in 13 hours on a single core to achieve a P@1 of 49.35\% whereas \alg could achieve 51.78\% by training 3 learners in 12 hours. Similarly, \alg's training time on Ads1M was 7 hours per learner on a single core. 

\alg's predictions could also be up to 300 times faster than LEML’s. For instance, on WikiLSHTC, \alg made predictions in 8 milliseconds per test point as compared to LEML's 279. \alg therefore brings the prediction time of embedding methods to be much closer to that of tree based methods (FastXML took 0.5 milliseconds per test point on WikiLSHTC) and within the acceptable limit of most real world applications.

\begin{table}[t!]
\label{tab:expResLS}
\renewcommand*{\arraystretch}{0.8}
\tabcolsep=0.11cm
\captionsetup{font=small}
    \caption{\textbf{Precision Accuracies} (a) Large-scale data sets : Our proposed method \alg is as much as 35\% more accurate in terms of P@1 and 22\% in terms of P@5 than LEML, a leading embedding method. Other embedding based methods do not scale to the large-scale data sets; we compare against them on small-scale data sets in Table \ref{tab:smallScale}. \alg is also  6\% more accurate (w.r.t. P@1 and P@5) than FXML, a state-of-the-art tree method. `-' indicates LEML could not be run with the standard resources. (b) Small-scale data sets : \alg consistently outperforms state of the art approaches. WSABIE, which also uses kNN classifier on its embeddings is significantly less accurate than \alg on all the data sets, showing the superiority of our embedding learning algorithm.\vspace*{-10pt}}    
    \begin{minipage}{.5\linewidth}
      \centering     
       
     \subfloat[]{\hspace*{-10pt}
	{\scriptsize
	\label{tab:expResLSa}
\begin{tabular}[h!]{@{}cc|ccccc@{}}
\toprule
Data set & & \alg & LEML & FastXML & LPSR-NB \\ \midrule
\multirow{3}{*}{Wiki10} & P@1 & \textbf{85.54} & 73.50 & 82.56 & 72.71 \\
& P@3 & \textbf{73.59} & 62.38 & 66.67 & 58.51 \\
& P@5 & \textbf{63.10} & 54.30 & 56.70 & 49.40 \\
\midrule
\multirow{3}{*}{Delicious-Large} & P@1 & \textbf{47.03} & 40.30 & 42.81 & 18.59 \\
& P@3 & \textbf{41.67} & 37.76 & 38.76 & 15.43 \\
& P@5 & \textbf{38.88} & 36.66 & 36.34 & 14.07 \\
\midrule
\multirow{3}{*}{WikiLSHTC} & P@1 & \textbf{55.57} & 19.82 & 49.35 & 27.91 \\
& P@3 & \textbf{33.84} & 11.43 & 32.69 & 16.04 \\
& P@5 & \textbf{24.07} & 8.39 & 24.03 & 11.57 \\
\midrule
\multirow{3}{*}{Amazon} & P@1 & \textbf{35.05} & 8.13 & 33.36 & 28.65 \\
& P@3 & \textbf{31.25} & 6.83 & 29.30 & 24.88 \\
& P@5 & \textbf{28.56}	 & 6.03 & 26.12 & 22.37 \\
\midrule
\multirow{3}{*}{Ads-1m} & P@1 & 21.84 & - & \textbf{23.11} & 17.08 \\
& P@3 & \textbf{14.30} & - & 13.86 & 11.38 \\
& P@5 & \textbf{11.01} & - & 10.12 & 8.83 \\
\bottomrule
\end{tabular}
}
   }
   \end{minipage} 
       \hspace*{2ex}
   \begin{minipage}{.5\linewidth}
         \centering
    \subfloat[]{\vspace{-2ex}
	{\scriptsize
	\label{tab:expResLSb}\hspace*{-20pt}
\begin{tabular}[h!]{@{}cc|cccccc@{}}
\toprule
Data set & & \alg & LEML & FastXML & WSABIE & OneVsAll\\ 
\midrule
\multirow{3}{*}{BibTex} & P@1 & \textbf{65.57} & 62.53  & 63.73 & 54.77 & 61.83 \\
& P@3 & \textbf{40.02}  & 38.4  & 39.00  & 32.38 & 36.44\\
& P@5 & \textbf{29.30} & 28.21 & 28.54 & 23.98 & 26.46\\ 
\midrule

\multirow{3}{*}{Delicious} & P@1 &  68.42 & 65.66 & \textbf{69.44} & 64.12 & 65.01 \\
& P@3 & 61.83 & 60.54 & \textbf{63.62} & 58.13 & 58.90\\
& P@5 & 56.80 & 56.08 & \textbf{59.10} & 53.64 & 53.26\\ 
\midrule

\multirow{3}{*}{MediaMill} & P@1 & \textbf{87.09} & 84.00 & 84.24 & 81.29 & 83.57\\
& P@3 & \textbf{72.44} & 67.19 & 67.39 & 64.74 & 65.50\\
& P@5 & \textbf{58.45} & 52.80 & 53.14 & 49.82 & 48.57\\ 
\midrule

\multirow{3}{*}{EurLEX} & P@1 & \textbf{80.17} & 61.28 & 68.69 & 70.87 & 74.96\\
& P@3 & \textbf{65.39} & 48.66 & 57.73 & 56.62 & 62.92\\
& P@5 & \textbf{53.75} & 39.91 & 48.00 & 46.2 & 53.42\\
\bottomrule
\end{tabular}
}
    }\hspace*{2ex}\vspace*{5.6ex}
    \end{minipage}
\vspace*{-15pt}
\end{table}

\textbf{Effect of clustering and multiple learners}: As mentioned in the introduction, other embedding methods could also be extended by clustering the data and then learning a local embedding in each cluster. Ensembles could also be learnt from multiple such clusterings. We extend LEML in such a fashion, and refer to it as LocalLEML,  by using exactly the same 300 clusters per learner in the ensemble as used in \alg for a fair comparison.  As can be seen in Figure~\ref{fig:learnerP1}, \alg significantly outperforms LocalLEML with a single \alg learner being much more accurate than an ensemble of even 10 LocalLEML learners. Figure~\ref{fig:learnerP1} also demonstrates that \alg's ensemble can be much more accurate at prediction as compared to the tree based FastXML ensemble (the same plot is also presented in the appendix depicting the variation in accuracy with model size in RAM rather than the number of learners in the ensemble). The figure also demonstrates that very few \alg learners need to be trained before  accuracy starts saturating. Finally, Table~\ref{tab:wikiStab} shows that the variance in \alg’s prediction accuracy (w.r.t. different cluster initializations) is very small, indicating that the method is stable even though clustering in more than a million dimensions. 

\textbf{Results on small data sets}: Table~\ref{tab:smallScale}, in the appendix, compares the performance of \alg to several popular methods including embeddings, trees, kNN and 1-vs-All SVMs. Even though the tail label problem is not acute on these data sets, and \alg was restricted to a single learner, \alg's predictions could be significantly more accurate than all the other methods (except on Delicious where \alg was ranked second). For instance, \alg could outperform the closest competitor on EurLex by 3\% in terms of P1. Particularly noteworthy is the observation that \alg  outperformed WSABIE~\cite{Weston11}, which performs kNN classification on linear embeddings, by as much as 10\% on multiple  data sets. This demonstrates the superiority of \alg's local distance preserving embeddings over the traditional low-rank embeddings.


\appendix

\section{Generalization Error Analysis}\label{app:genbound}
To present our results, we first introduce some notation: for any embedding matrix $A$ and dataset $\D$, let
\begin{align*}
\hat\L(A;\D) &:= \frac{1}{n(n-1)}\sum_{i=1}^n\sum_{j\neq i}\ell(A;(\x_i,\y_i),(\x_j,\y_j))\\
\tilde\L(A;\D) &:= \frac{1}{n}\sum_{i=1}^n\Enob{(\x,\y) \sim \P}{\ell(A;(\x,\y),(\x_i,\y_i))}\\
\L(A) &:= \Enob{(\x,\y), (\tilde{\x},\tilde{\y})\sim \P}{\ell(A;(\x,\y),(\tilde{\x},\tilde{\y}))}\\
\end{align*}
We assume, without loss of generality that the data points are confined to a unit ball i.e. $\norm{\x}_2 \leq 1$ for all $\x\in\X$. Also let $Q = C\cdot\br{\lbar (r +\lbar)}$ where $\lbar$ is the average number of labels active in a data point, $r = \frac{\lbar}{\lambda}$, $\lambda$ and $\mu$ are the regularization constants used in \eqref{eq:nl2}, and $C$ is a {\em universal constant}.

\begin{repthm}{thm:plain-excess-risk-bd}
Assume that all data points are confined to a ball of radius $R$ i.e $\norm{x}_2 \leq R$ for all $x\in\X$. Then with probability at least $1 - \delta$ over the sampling of the data set $\D$, the solution $\hat A$ to the optimization problem \eqref{eq:nl2} satisfies,
\[
\L(\hat A) \leq \inf_{A^\ast\in\A}8\bc{\L(A^\ast) + C\br{\lbar^2 + \br{r^2+\norms{A^\ast}_F^2}R^4}\sqrt{\frac{1}{n}\log\frac{1}{\delta}}},
\]
where $r = \frac{\lbar}{\lambda}$, and $C$ and $C'$ are universal constants.
\end{repthm}
\begin{proof}
Our proof will proceed in the following steps. Let $A^\ast$ be the population minimizer of the objective in the statement of the theorem.
\begin{enumerate}
	\item \textbf{Step 1} (Capacity bound): we will show that for some $r$, we have $\norms{\hat A}_F \leq r$
	\item \textbf{Step 2} (Uniform convergence): we will show that w.h.p., $\sup_{\substack{A\in\A\\\norm{A}\leq r}}\bc{\L(A) - \hat\L(A;\D)} \leq \O{\sqrt{\frac{1}{n}\log\frac{1}{\delta}}}$
	\item \textbf{Step 3} (Point convergence): we will show that w.h.p., ${\hat\L(A^\ast;\D) - \L(A^\ast)} \leq \O{\sqrt{\frac{1}{n}\log\frac{1}{\delta}}}$
\end{enumerate}
Having these results will allow us to prove the theorem in the following manner
\begin{align*}
\L(\hat A) &\leq \hat\L(\hat A,\D) + \sup_{\substack{A\in\A\\\norm{A}\leq r}}\bc{\hat\L(A;\D) - \L(A)}\leq \hat\L(A^\ast,\D) + \O{\sqrt{\frac{1}{n}\log\frac{1}{\delta}}}
		   \leq \L(A^\ast) + \O{\sqrt{\frac{1}{n}\log\frac{1}{\delta}}},
\end{align*}
where the second step follows from the fact that $\hat A$ is the empirical risk minimizer.

We will now prove these individual steps as separate lemmata, where we will also reveal the exact constants in these results.

\begin{lem}[Capacity bound]
For the regularization parameters chosen for the loss function $\ell(\cdot)$, the following holds for the minimizer $\hat A$ of \eqref{eq:nl2}
\[
\norms{\hat A}_F \leq Tr(A)\leq \frac{1}{\lambda} \bar{L}.
\]
\end{lem}
\begin{proof}
Since, $\hat A$ minimizes \eqref{eq:nl2}, we have: 
$$\|A\|_F\leq Tr(A)\leq \frac{1}{\lambda} \frac{1}{n(n-1)}\sum_{ij}(\ip{\y_i}{\y_j}^2\leq \frac{1}{\lambda} \max_{ij} \ip{\y_i}{\y_j}.$$ 
\end{proof}
The above result shows that we can, for future analysis, restrict our hypothesis space to
\[
\widetilde\A(r) := \bc{A \in \A: \norm{A}_F^2 \leq r^2},
\]
where we set $r = \frac{\lbar}{\lambda}$. This will be used to prove the following result.
\begin{lem}[Uniform convergence]
With probability at least $1 -\delta$ over the choice of the data set $\D$, we have
\[
{\hat\L(\hat A;\D) - \L(\hat A)} \leq 6\br{rR^2 + \lbar}^2\sqrt{\frac{1}{2n}\log\frac{1}{\delta}}
\]
\end{lem}
\begin{proof}
For notional simplicity, we will denote a labeled sample as $\z = (\x,\y)$. Given any two points $\z,\z' \in\ZZ = \X\times\Y$ and any $A\in\widetilde\A(r)$, we will then write
\[
\ell(A;\z,\z') = g(\ip{{\y}}{\y'})\br{\ip{{\y}}{\y'}-{\x}^T A\x'}^2 + \lambda Tr(A),
\]
so that, for the training set $\D = \bc{\z_1,\ldots,\z_n}$, we have
\[
\hat\L(A;\D) = \frac{1}{n(n-1)}\sum_{i=1}^n\sum_{j\neq i}\ell(A;\z_i,\z_j)
\\
\]
as well as
\[
\L(A) = \Enob{z,z'\sim \P}{\ell(A;\z,\z')}.
\]
Note that we ignore the $\|V\x\|_1$ term in \eqref{eq:nl2} completely because it is a regularization term and it won't increase the excess risk. 

Suppose we draw a fresh data set $\widetilde\D = \bc{\tz_1,\ldots,\tz_n} \sim \P$, then we have, by linearity of expectation,
\[
\Enob{\widetilde\D\sim\P}{\hat\L(\hat A;\tilde\D)} = \frac{1}{n(n-1)}\sum_{i=1}^n\sum_{j\neq i}\Enob{\widetilde\D\sim\P}{\ell(A;\tz_i,\tz_j)} = \L(\hat A).
\]
Now notice that for any $A\in\widetilde\A$, suppose we perturb the data set $\D$ at the $i\th$ location to get a perturbed data set $\D^i$, then the following holds
\[
\abs{\hat\L(A;\D) - \hat\L(A;\D^i)} \leq \frac{4(\lbar^2 + r^2R^4)}{n}.
\]
which allows us to bound the excess risk as follows
\begin{align*}
\L(\hat A) - \hat\L(\hat A;\D) &= \Enob{\widetilde\D\sim\P}{\hat\L(\hat A;\tilde\D)} - \hat\L(\hat A;\D)							   \leq \sup_{A\in\widetilde\A(r)}\bc{\Enob{\widetilde\D\sim\P}{\hat\L( A;\tilde\D)} - \hat\L( A;\D)}\\
							   &\leq \Enob{\D\sim\P}\sup_{A\in\widetilde\A(r)}\bc{\Enob{\widetilde\D\sim\P}{\hat\L( A;\tilde\D)} - \hat\L( A;\D)} + {4(\lbar^2 + r^2R^4)}\sqrt{\frac{1}{2n}\log\frac{1}{\delta}}\\
							   &\leq \underbrace{\Enob{\D,\widetilde\D\sim\P}\sup_{A\in\widetilde\A(r)}\bc{{\hat\L( A;\tilde\D)} - \hat\L( A;\D)}}_{Q_n(\widetilde\A(r))} + {4(\lbar^2 + r^2R^4)}\sqrt{\frac{1}{2n}\log\frac{1}{\delta}},
\end{align*}
where the third step follows from an application of McDiarmid's inequality and the last step follows from Jensen's inequality. We now bound the quantity $Q_n(\widetilde\A(r))$ below. Let $\bar\ell(A,\z,\z') := \ell(A,\z,\z') - \lambda\cdot Tr(A)$. Then we have
\begin{align*}
Q_n(\widetilde\A(r)) &= \Enob{\D,\widetilde\D\sim\P}\sup_{A\in\widetilde\A(r)}\bc{{\hat\L( A;\tilde\D)} - \hat\L( A;\D)}\\
					 &= \frac{1}{n(n-1)}\EE{\z_i,\tz_i\sim\P}{\sup_{A\in\widetilde\A(r)}\bc{\sum_{i=1}^n\sum_{j\neq i}\ell(A;\tz_i,\tz_j) - \ell(A;\z_i,\z_j)}}\\
					 &= \frac{1}{n(n-1)}\EE{\z_i,\tz_i\sim\P}{\sup_{A\in\widetilde\A(r)}\bc{\sum_{i=1}^n\sum_{j\neq i}\bar\ell(A;\tz_i,\tz_j) - \bar\ell(A;\z_i,\z_j)}}\\
					 &\leq \frac{2}{n}\EE{\z_i,\tz_i}{\sup_{A\in\widetilde\A(r)}\bc{\sum_{i=1}^{n/2}\bar\ell(A;\tz_i,\tz_{n/2+i}) - \bar\ell(A;\z_i,\z_{n/2+i})}}\\
					 &\leq 2\cdot\underbrace{\frac{2}{n}\EE{z_i,\epsilon_i}{\sup_{A\in\widetilde\A(r)}\bc{\sum_{i=1}^{n/2}\epsilon_i\bar\ell(A;\z_i,\z_{n/2+i})}}}_{\RR_n(\ell\circ\widetilde\A(r))}= 2\cdot\RR_{n/2}(\ell\circ\widetilde\A(r))
\end{align*}
where the last step uses a standard symmetrization argument with the introduction of the Rademacher variables $\epsilon_i \sim {-1,+1}$. The second step presents a stumbling block in the analysis since the interaction between the pairs of the points means that traditional symmetrization can no longer done. Previous works analyzing such ``pairwise'' loss functions face similar problems \cite{KarSJK13}. Consequently, this step uses a powerful alternate representation for U-statistics to simplify the expression. This technique is attributed to Hoeffding. This, along with the Hoeffding decomposition, are two of the most powerful techniques to deal with ``coupled'' random variables as we have in this situation.
\begin{thm}
For any set of real valued functions $q_\tau : \X \times \X \rightarrow \R$ indexed by $\tau \in T$, if $X_1,\ldots,X_n$ are i.i.d. random variables then we have
\[
\E{\underset{\tau \in T}{\sup} \frac{2}{n(n-1)}\sum_{1\leq i < j \leq n} q_\tau(X_i,X_j)} \leq \E{\underset{\tau \in T}{\sup} \frac{2}{n}\sum_{i = 1}^{n/2} q_\tau(X_i,X_{n/2 + i})}
\]
\end{thm}
Applying this decoupling result to the random variables $X_i = (\tz_i,\z_i)$, the index set $\widetilde\A(r)$ and functions $q_A(X_i,X_j) = \ell(A;\tz_i,\tz_j) - \ell(A;\z_i,\z_j) = \bar\ell(A;\tz_i,\tz_j) - \bar\ell(A;\z_i,\z_j)$ gives us the second step. We now concentrate on bounding the resulting Rademacher average term $\RR_n(\ell\circ\widetilde\A(r))$. We have
\begin{align*}
\RR_{n/2}(\ell\circ\widetilde\A(r)) ={}& \frac{2}{n}\EE{z_i,\epsilon_i}{\sup_{A\in\widetilde\A(r)}\bc{\sum_{i=1}^{n/2}\epsilon_i\bar\ell(A;\z_i,\z_{n/2+i})}}\\
								={}& \frac{2}{n}\EE{z_i,\epsilon_i}{\sup_{A\in\widetilde\A(r)}\bc{\sum_{i=1}^{n/2}\epsilon_ig(\ip{\y_i}{\y_{n/2+i}})\br{\ip{\y_i}{\y_{n/2+i}}-\x_i^T A\x_{n/2+i}}^2}}. \end{align*}
That is,
\begin{align*}
\RR_{n/2}(\ell\circ\widetilde\A(r))\leq{}& \underbrace{\frac{2}{n}\EE{z_i,\epsilon_i}{{\sum_{i=1}^{n/2}\epsilon_ig(\ip{\y_i}{\y_{n/2+i}})\ip{\y_i}{\y_{n/2+i}}^2}}}_{(A)}\\
								&+ \underbrace{\frac{2}{n}\EE{z_i,\epsilon_i}{\sup_{A\in\widetilde\A(r)}\bc{\sum_{i=1}^{n/2}\epsilon_ig(\ip{\y_i}{\y_{n/2+i}})\br{\x_i^T A\x_{n/2+i}}^2}}}_{B_n(\ell\circ\widetilde\A(r))}\\
								&+ \underbrace{\frac{4}{n}\EE{z_i,\epsilon_i}{\sup_{A\in\widetilde\A(r)}\bc{\sum_{i=1}^{n/2}\epsilon_ig(\ip{\y_i}{\y_{n/2+i}})\ip{\y_i}{\y_{n/2+i}}\br{\x_i^T A\x_{n/2+i}}}}}_{C_n(\ell\circ\widetilde\A(r))}
\end{align*}
Now since the random variables $\epsilon_i$ are zero mean and independent of $z_i$, we have $\Enob{\epsilon_i | z_i, z_{n/2+i}}\epsilon_i = 0$ which we can use to show that $\EE{\epsilon_i | z_i, z_{n/2+i}}{\epsilon_ig(\ip{\y_i}{\y_{n/2+i}})\ip{\y_i}{\y_{n/2+i}}^2} = 0$ which gives us, by linearity of expectation, $(A) = 0$. To bound the next two terms we use the following standard contraction inequality: 
\begin{thm}
Let $\H$ be a set of bounded real valued functions from some domain $\X$ and let $\vecx_1,\ldots,\vecx_n$ be arbitrary elements from $\X$. Furthermore, let $\phi_i : \R \rightarrow \R$, $i = 1,\ldots,n$ be $L$-Lipschitz functions such that $\phi_i(0) = 0$ for all $i$. Then we have
\[
\E{\underset{h \in \H}{\sup}\frac{1}{n}\sum_{i=1}^n \epsilon_i\phi_i(h(\vecx_i))} \leq L \E{\underset{h \in \H}{\sup}\frac{1}{n}\sum_{i=1}^n \epsilon_ih(\vecx_i)}.
\]
\end{thm}
Now define
\[
\phi_i(w) = g(\ip{\y_i}{\y_{n/2+i}}){w}^2
\]
Clearly $\phi_i(0) = 0$ and $0 \leq g(\ip{\y_i}{\y_{n/2+i}}) \leq 1 $. Moreover, in our case $w = x^TAx'$ for some $A \in \widetilde\A(r)$ and $\norm{x},\norm{x'} \leq R$. Thus, the function $\phi_i(\cdot)$ is $rR^2$-Lipschitz. Note that here we exploit the fact that the contraction inequality is actually proven for the empirical Rademacher averages due to which we can take  $g(\ip{\y_i}{\y_{n/2+i}})$ to be a constant dependent only on $i$. This allows us to bound the term $B_n(\ell\circ\widetilde\A(r))$ as follows
\begin{align*}
B_n(\ell\circ\widetilde\A(r)) &= \frac{2}{n}\EE{z_i,\epsilon_i}{\sup_{A\in\widetilde\A(r)}\bc{\sum_{i=1}^{n/2}\epsilon_ig(\ip{\y_i}{\y_{n/2+i}})\br{\x_i^T A\x_{n/2+i}}^2}}\\
							  &\leq rR^2\cdot\underbrace{\frac{2}{n}\EE{z_i,\epsilon_i}{\sup_{A\in\widetilde\A(r)}\bc{\sum_{i=1}^{n/2}\epsilon_i\br{\x_i^T A\x_{n/2+i}}}}}_{\RR_{n/2}(\widetilde\A(r))}\leq rR^2\cdot\RR_{n/2}(\widetilde\A(r)).
\end{align*}
Similarly, we can show that
\begin{align*}
C_n(\ell\circ\widetilde\A(r)) &= \frac{4}{n}\EE{z_i,\epsilon_i}{\sup_{A\in\widetilde\A(r)}\bc{\sum_{i=1}^{n/2}\epsilon_ig(\ip{\y_i}{\y_{n/2+i}})\ip{\y_i}{\y_{n/2+i}}\br{\x_i^T A\x_{n/2+i}}}}\\
							  &\leq \frac{4\lbar}{n}\EE{z_i,\epsilon_i}{\sup_{A\in\widetilde\A(r)}\bc{\sum_{i=1}^{n/2}\epsilon_i\br{\x_i^T A\x_{n/2+i}}}}\\
							  &\leq  2\lbar\cdot\RR_{n/2}(\widetilde\A(r)).
\end{align*}
Thus, we have
\[
\RR_{n/2}(\ell\circ\widetilde\A(r)) \leq \br{rR^2 + 2\lbar}\cdot\RR_{n/2}(\widetilde\A(r))
\]
Now all that remains to be done is bound $\RR_n(\ell\circ\widetilde\A(r))$. This can be done by invoking standard bounds on Rademacher averages for regularized function classes. In particular, using the two stage proof technique outlined in \cite{KarSJK13}, we can show that
\[
\RR_{n/2}(\widetilde\A(r)) \leq rR^2\sqrt\frac{2}{n}
\]
Putting it all together gives us the following bound: with probability at least $1 - \delta$, we have
\[
\L(\hat A) - \hat\L(\hat A;\D) \leq 2(rR^2 + 2\lbar)rR^2\sqrt\frac{2}{n} + {4(\lbar^2 + r^2R^4)}\sqrt{\frac{1}{2n}\log\frac{1}{\delta}}
\]
as claimed
\end{proof}
The final part shows pointwise convergence for the population risk minimizer.
\begin{lem}[Point convergence]
With probability at least $1 -\delta$ over the choice of the data set $\D$, we have
\[
{\hat\L(A^\ast;\D) - \L(A^\ast)} \leq {4(\lbar^2 + \norms{A^\ast}_F^2R^4)}\sqrt{\frac{1}{2n}\log\frac{1}{\delta}},
\]
where $A^\ast$ is the population minimizer of the objective in the theorem statement.
\end{lem}
\begin{proof}
We note that, as before
\[
\Enob{\D\sim\P}{\hat\L(A^\ast,\D)} = \L(A^\ast)
\]
Let $\D$ be a realization of the sample and $\D^i$ be a perturbed data set where the $i^\th$ data point is arbitrarily perturbed. Then we have
\[
\abs{\hat\L(A^\ast;\D) - \hat\L(A^\ast;\D^i)} \leq \frac{4\br{\lbar^2 + \norms{A^\ast}_F^2R^4}}{n}.
\]

Thus, an application of McDiarmid's inequality shows us that with probability at least $1 -\delta$, we have
\[
{\hat\L(A^\ast;\D) - \L(A^\ast)} = {\hat\L(A^\ast;\D) - \Enob{\D\sim\P}{\hat\L(A^\ast;\D)}} \leq {4\br{\lbar^2 + \norms{A^\ast}_F^2R^4}}\sqrt{\frac{1}{2n}\log\frac{1}{\delta}},
\]
which proves the claim.
\end{proof}
Putting the three lemmata together as shown above concludes the proof of the theorem.
\end{proof}

Although the above result ensures that the embedding provided by $\hat A$ would preserve neighbors over the population, in practice, we are more interested in preserving the neighbors of test points among the training points, as they are used to predict the label vector. The following extension of our result shows that $\hat A$ indeed accomplishes this as well.
\begin{thm}\label{thm:excess-risk-to-train-bd}
Assume that all data points are confined to a ball of radius $R$ i.e $\norm{x}_2 \leq R$ for all $x\in\X$. Then with probability at least $1 - \delta$ over the sampling of the data set $\D$, the solution $\hat A$ to the optimization problem \eqref{eq:nl2} ensures that,
\[
\tilde\L(\hat A;\D) \leq \inf_{A^\ast\in\A}\bc{\tilde\L(A^\ast;\D) + C\br{\lbar^2 + \br{r^2+\norms{A^\ast}_F^2}R^4}\sqrt{\frac{1}{n}\log\frac{1}{\delta}}},
\]
where $r = \frac{\lbar}{\lambda}$, and $C$ is a universal constant.
\end{thm}
Note that the loss function $\tilde\L(A;\D)$ exactly captures the notion of how well an embedding matrix $A$ can preserve the neighbors of an unseen point among the training points.
\begin{proof}
We first recall and rewrite the form of the loss function considered here. For any data set $\D = \bc{z_1,\ldots,z_n}$. For any $A\in\A$ and $z \in \Z$, let $\wp(A;z) := \Enob{z' \sim \P}{\ell(A;z,z')}$. This allows us to write
\[
\tilde\L(A;\D) := \frac{1}{n}\sum_{i=1}^n\Enob{z \sim \P}{\ell(A;z,z_i)} = \frac{1}{n}\sum_{i=1}^n{\wp(A;z_i)}
\]
Also note that for any fixed $A$, we have
\[
\Enob{\D\sim\P}{\tilde\L(A;\D)} = \L(A).
\]
Now, given a perturbed data set $\D^i$, we have
\[
\abs{\tilde\L(A;\D) - \tilde\L(A;\D^i)} \leq \frac{4(\lbar^2 + r^2R^4)}{n},
\]
as before. Since this problem does not have to take care of pairwise interactions between the data points (since the ``other'' data point is being taken expectations over), using standard Rademacher style analysis gives us, with probability at least $1 - \delta$,
\[
{\tilde\L(\hat A;\D) - \L(\hat A)} \leq 2\br{rR^2 + 2\lbar}rR^2\sqrt\frac{2}{n} + 4(\lbar^2 + r^2R^4)\sqrt{\frac{1}{2n}\log\frac{1}{\delta}}
\]
A similar analysis also gives us with the same confidence
\[
{\L(A^\ast) - \hat\L(A^\ast;\D)} \leq 4(\lbar^2 + \norms{A^\ast}_F^2R^4)\sqrt{\frac{1}{2n}\log\frac{1}{\delta}}
\]
However, an argument similar to that used in the proof of Theorem~\ref{thm:plain-excess-risk-bd} shows us that
\[
\L(\hat A) \leq {\L(A^\ast) + C\br{\lbar^2 + \br{r^2+\norms{A^\ast}_F^2}R^4}\sqrt{\frac{1}{n}\log\frac{1}{\delta}}}
\]
Combining the above inequalities yields the desired result.
\end{proof}



\newpage
\section{Experiments}
\label{app:results}

\begin{table}[th]
  \newcommand{\tabcolwidth}[1]{{\centering \mbox{#1}}}
  \centering
  \small{
  \caption{Data set Statistics: $n$ and $m$ are the number of training and test points respectively, $d$ and $L$ are the number of features and labels, respectively, and $\bar{d}$ and $\bar{L}$ are the average number of nonzero features and positive labels in an instance, respectively.}
  \label{tab:stats}\vspace{1ex}
 {
	 
	  \begin{tabular}{@{}lcccccccc@{}}
	  \toprule 
	  \multicolumn{1}{c}{\multirow{1}{*}{Data set}} & \multicolumn{1}{c}{\multirow{1}{*}{\textit{d}}} & \multicolumn{1}{c}{\multirow{1}{*}{\textit{L}}} & \multicolumn{1}{c}{\multirow{1}{*}{\textit{n}}} & \multicolumn{1}{c}{\multirow{1}{*}{\textit{m}}} & \multicolumn{1}{c}{\multirow{1}{*}{\textit{$\bar{d}$}}} & \multicolumn{1}{c}{\multirow{1}{*}{\textit{$\bar{L}$}}}\\
      \midrule
MediaMill	&	120	&	101	&	30993	&	12914	&	120.00	&	4.38	\\
BibTeX	&	1836	&	159	&	4880	&	2515	&	68.74	&	2.40	\\
Delicious	&	500	&	983	&	12920	&	3185	&	18.17	&	19.03	\\
EURLex	&	5000	&	3993	&	17413	&	1935	&	236.69	&	5.31	\\
Wiki10	&	101938	&	30938	&	14146	&	6616	&	673.45	&	18.64	\\
DeliciousLarge	&	782585	&	205443	&	196606	&	100095	&	301.17	&	75.54	\\
WikiLSHTC	&	1617899	&	325056	&	1778351	&	587084	&	42.15	&	3.19	\\
Amazon	&	135909	&	670091	&	490449	&	153025	&	75.68	&	5.45	\\
Ads1M	&	164592	&	1082898	&	3917928	&	1563137	&	9.01	&	1.96	\\
      \bottomrule
    \end{tabular}
    }
	}
\end{table}

\begin{table*}[h!]

\captionsetup{font=small}
\centering
\caption{\small{Results on Small Scale data sets : Comparison of precision accuracies of \alg with competing baseline methods on small scale data sets. The results reported are average precision values along with standard deviations over 10 random train-test split for each Data set. \alg outperforms all baseline methods on all data sets (except Delicious, where it is ranked $2^{nd}$ after FastXML)}}
 \label{tab:smallScale}
\tiny{
\tabcolsep=0.11cm
\begin{tabular}{@{}cccccccccccccccc@{}}\toprule
\multicolumn{2}{c}{Data set} & \multicolumn{1}{c}{Proposed} & \phantom{abc} & \multicolumn{5}{c}{Embedding} &
\phantom{abc} & \multicolumn{3}{c}{Tree Based}  & \phantom{abc} & \multicolumn{2}{c}{Other}\\
\cmidrule{5-9} \cmidrule{11-13} \cmidrule{15-16}
 &  & \alg && LEML & WSABIE  & CPLST & CS & ML-CSSP && FastXML-1 & FastXML & LPSR && OneVsAll & KNN \\ \midrule
\multirow{3}{*}{Bibtex} & P@1 & \textbf {65.57 \textpm 0.65} && 62.53\textpm 0.69  & 54.77\textpm 0.68 & 62.38 \textpm 0.42 & 58.87 \textpm 0.64 & 44.98 \textpm 0.08 && 37.62 \textpm 0.91 & 63.73\textpm 0.67 & 62.09\textpm 0.73 && 61.83 \textpm 0.77 & 57.00 \textpm 0.85\\
& P@3 & \textbf{40.02 \textpm 0.39} && 38.40 \textpm 0.47  & 32.38 \textpm 0.26 & 37.83 \textpm 0.52 & 33.53 \textpm 0.44 & 30.42 \textpm 2.37 && 24.62 \textpm 0.68 & 39.00 \textpm 0.57 & 36.69 \textpm 0.49 && 36.44 \textpm 0.38 & 36.32 \textpm 0.47\\
& P@5 & \textbf{29.30 \textpm 0.32} && 28.21 \textpm 0.29  & 23.98 \textpm 0.18 & 27.62 \textpm 0.28 & 23.72 \textpm 0.28 & 23.53 \textpm 1.21 && 21.92 \textpm 0.65 & 28.54 \textpm 0.38 & 26.58 \textpm 0.38 && 26.46 \textpm 0.26 & 28.12 \textpm 0.39 \\
\midrule
\multirow{3}{*}{Delicious} & P@1 & 68.42 \textpm 0.53 && 65.66 \textpm 0.97  & 64.12 \textpm 0.77 & 65.31 \textpm 0.79 & 61.35 \textpm 0.77 & 63.03 \textpm 1.10 && 55.34 \textpm 0.92 & \textbf{69.44 \textpm 0.58} & 65.00\textpm 0.77 && 65.01 \textpm 0.73 & 64.95 \textpm 0.68\\
& P@3 & 61.83 \textpm 0.59 && 60.54 \textpm 0.44  & 58.13 \textpm 0.58 &59.84 \textpm 0.5 & 56.45 \textpm 0.62 & 56.26 \textpm 1.18 && 50.69 \textpm 0.58 & \textbf{63.62 \textpm 0.75} & 58.97 \textpm 0.65 && 58.90 \textpm 0.60 & 58.90 \textpm 0.70\\
& P@5 & 56.80 \textpm 0.54 && 56.08 \textpm 0.56  & 53.64 \textpm 0.55 & 55.31 \textpm 0.52 & 52.06 \textpm 0.58 & 50.15 \textpm 1.57 && 45.99 \textpm 0.37 & \textbf{59.10 \textpm 0.65} & 53.46 \textpm 0.46 && 53.26 \textpm 0.57 & 54.12 \textpm 0.57\\
\midrule
\multirow{3}{*}{MediaMill} & P@1 & \textbf{87.09 \textpm 0.33} && 84.00\textpm 0.30  & 81.29 \textpm 1.70 & 83.34 \textpm 0.45 & 83.82 \textpm 0.36 & 78.94 \textpm 10.1 && 61.14\textpm 0.49 & 84.24 \textpm 0.27 & 83.57 \textpm 0.26 && 83.57 \textpm 0.25 & 83.46 \textpm 0.19\\
& P@3 & \textbf{72.44 \textpm 0.30} && 67.19 \textpm 0.29  & 64.74 \textpm 0.67 & 66.17 \textpm 0.39 & 67.31 \textpm 0.17 & 60.93 \textpm 8.5 && 53.37 \textpm 0.30 & 67.39 \textpm 0.20 & 65.78 \textpm 0.22 && 65.50 \textpm 0.23 & 67.91 \textpm 0.23\\
& P@5 & \textbf{58.45 \textpm 0.34} && 52.80 \textpm 0.17  & 49.82 \textpm 0.71 & 51.45 \textpm 0.37 & 52.80 \textpm 0.18 & 44.27 \textpm 4.8 && 48.39 \textpm 0.19 & 53.14 \textpm 0.18 & 49.97 \textpm 0.48 && 48.57 \textpm 0.56 & 54.24 \textpm 0.21\\
\midrule
\multirow{3}{*}{EurLEX} & P@1 & \textbf{80.17 \textpm 0.86} && 61.28\textpm 1.33  & 70.87 \textpm 1.11 & 69.93\textpm 0.90 & 60.18 \textpm 1.70 & 56.84\textpm 1.5 && 49.18 \textpm 0.55 & 68.69 \textpm 1.63 & 73.01 \textpm 1.4 && 74.96 \textpm 1.04 & 77.2 \textpm 0.79\\
& P@3 & \textbf{65.39 \textpm 0.88} && 48.66 \textpm 0.74  & 56.62 \textpm 0.67 & 56.18 \textpm 0.66 & 48.01 \textpm 1.90 & 45.4 \textpm 0.94 && 42.72 \textpm 0.51 & 57.73 \textpm 1.58 & 60.36 \textpm 0.56 && 62.92 \textpm 0.53 & 61.46 \textpm 0.96\\
& P@5 & \textbf{53.75 \textpm 0.80} && 39.91 \textpm 0.68  & 46.20 \textpm 0.55 & 45.74 \textpm 0.42 & 38.46 \textpm 1.48 & 35.84 \textpm 0.74 && 37.35 \textpm 0.42 & 48.00 \textpm 1.40 & 50.46 \textpm 0.50 && 53.42 \textpm 0.37 & 50.45 \textpm 0.64\\
\bottomrule
\end{tabular}

}
\end{table*}

\begin{table*}[h!]

\captionsetup{font=small} 
\centering
\tiny{
\tabcolsep=0.11cm
\caption{\small{Stability of \alg learners. We show mean precision values over 10 runs of \alg on WikiLSHTC with varying number of learners. Each individual learner as well as ensemble of \alg learners was found to be extremely stable with with standard deviation ranging from 0.16\% on P1 to 0.11\% on P5.}}
\label{tab:wikiStab}
\begin{tabular}{@{}ccccccccccc@{}}
\toprule
$\#$ Learners & 1 & 2 & 3 & 4 & 5 & 6 & 7 & 8 & 9 & 10 \\
\midrule
P@1 & 46.04 \textpm 0.1659 & 50.04 \textpm 0.0662  & 51.65 \textpm 0.074 & 52.62 \textpm 0.0878 & 53.28 \textpm 0.0379 & 53.63 \textpm 0.083 & 54.03 \textpm 0.0757 & 54.28 \textpm 0.0699 & 54.44 \textpm 0.048 & 54.69 \textpm 0.035 \\
P@3 & 26.15 \textpm 0.1359 & 29.32 \textpm 0.0638  & 30.70 \textpm 0.052 & 31.55 \textpm 0.067 & 32.14 \textpm 0.0351 & 32.48 \textpm 0.0728 & 32.82 \textpm 0.0694 & 33.07 \textpm 0.0503 & 33.24 \textpm 0.023 & 33.45 \textpm 0.0127 \\
P@5 & 18.14 \textpm 0.1045 & 20.58 \textpm 0.0517  & 21.68 \textpm 0.0398 & 22.36 \textpm 0.0501 & 22.85 \textpm 0.0179 & 23.12 \textpm 0.0525 & 23.4 \textpm 0.0531 & 23.60 \textpm 0.0369 & 23.74 \textpm 0.0172 & 23.92 \textpm 0.0115 \\

\bottomrule
\end{tabular}
}
\end{table*}

\begin{figure}[h!]
\captionsetup{font=small} 
  \centering
\begin{tabular}{ccc}
  \includegraphics[width=.33\textwidth]{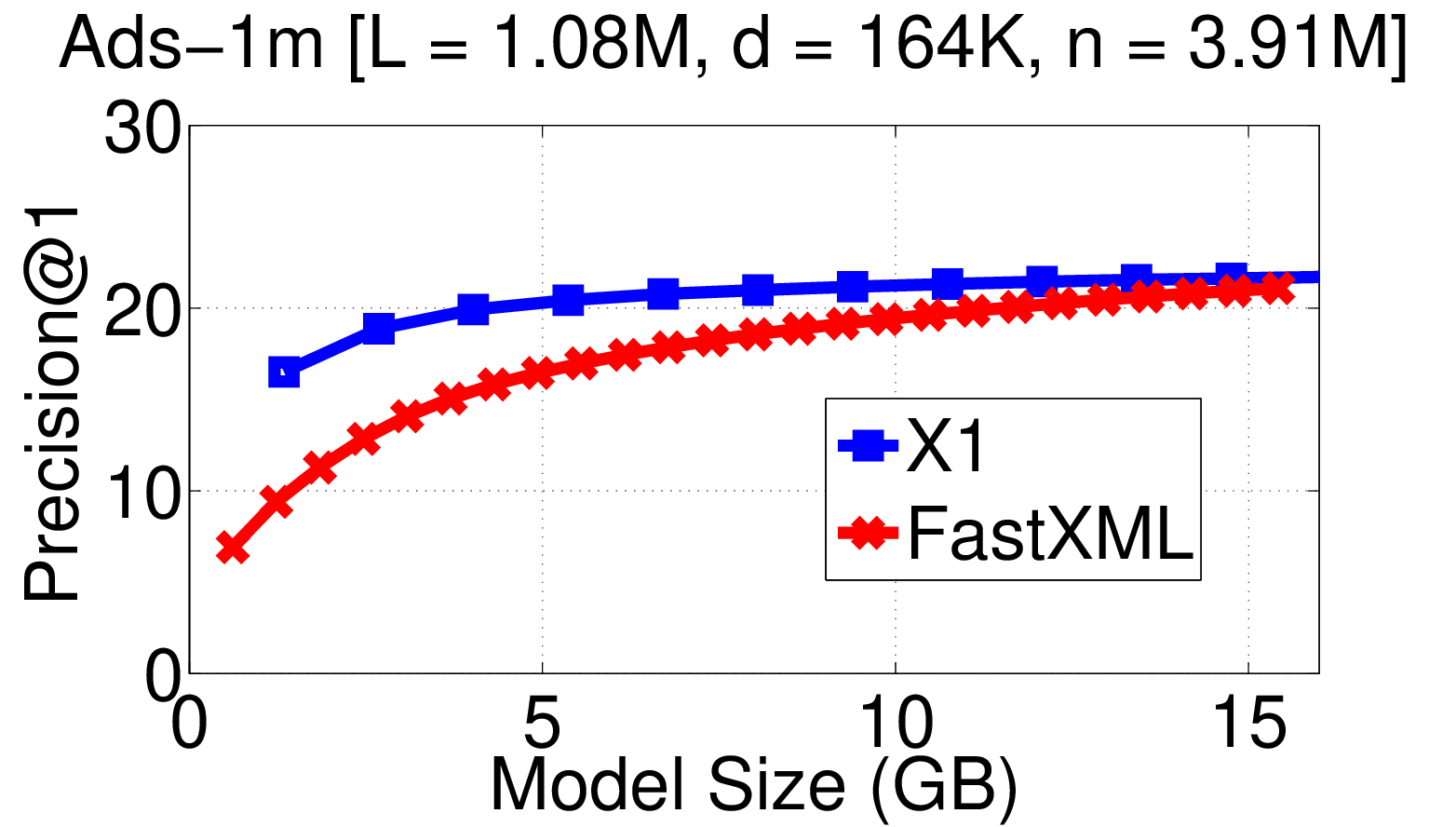}&
  \includegraphics[width=.33\textwidth]{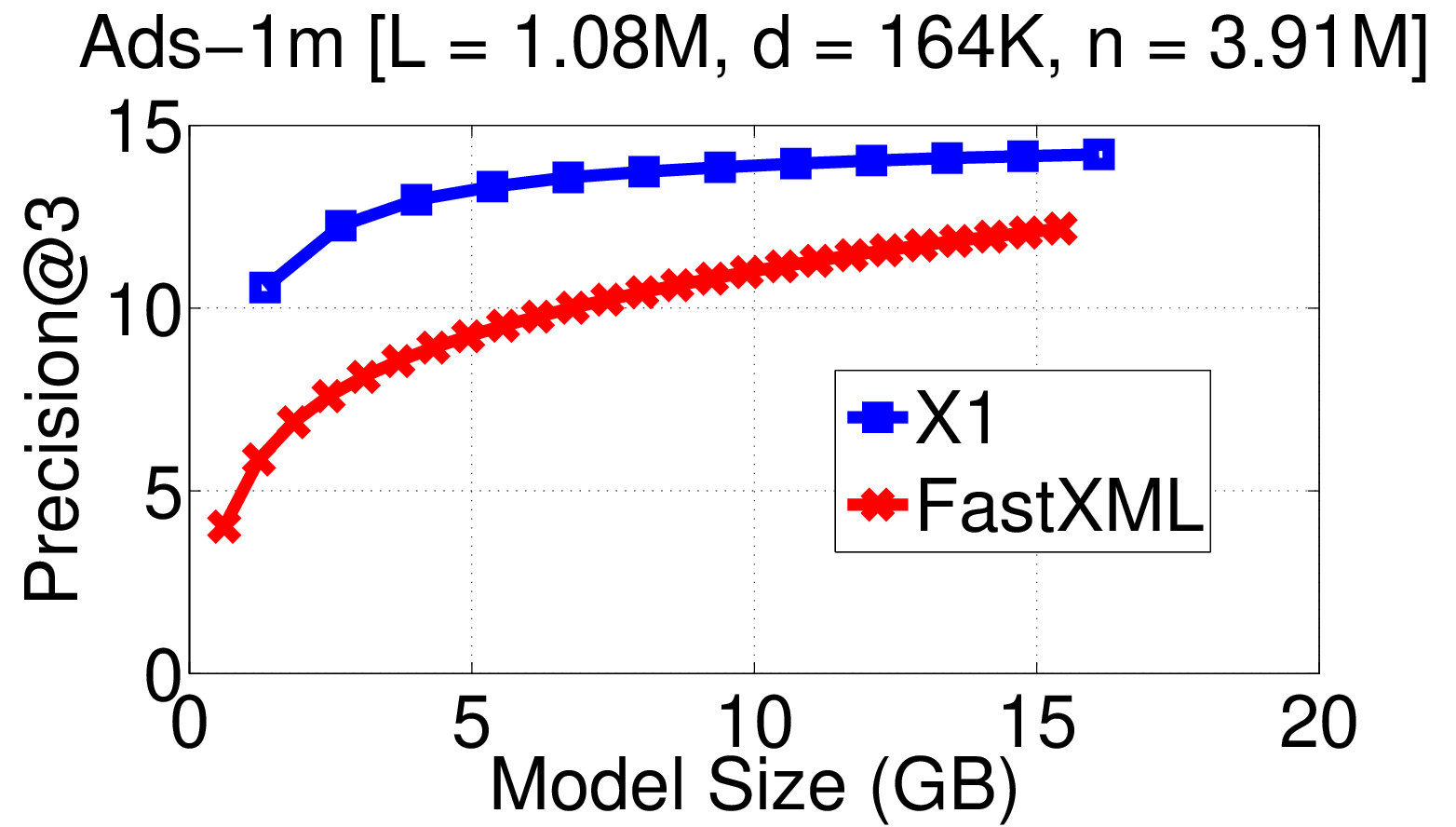}&
  \includegraphics[width=.33\textwidth]{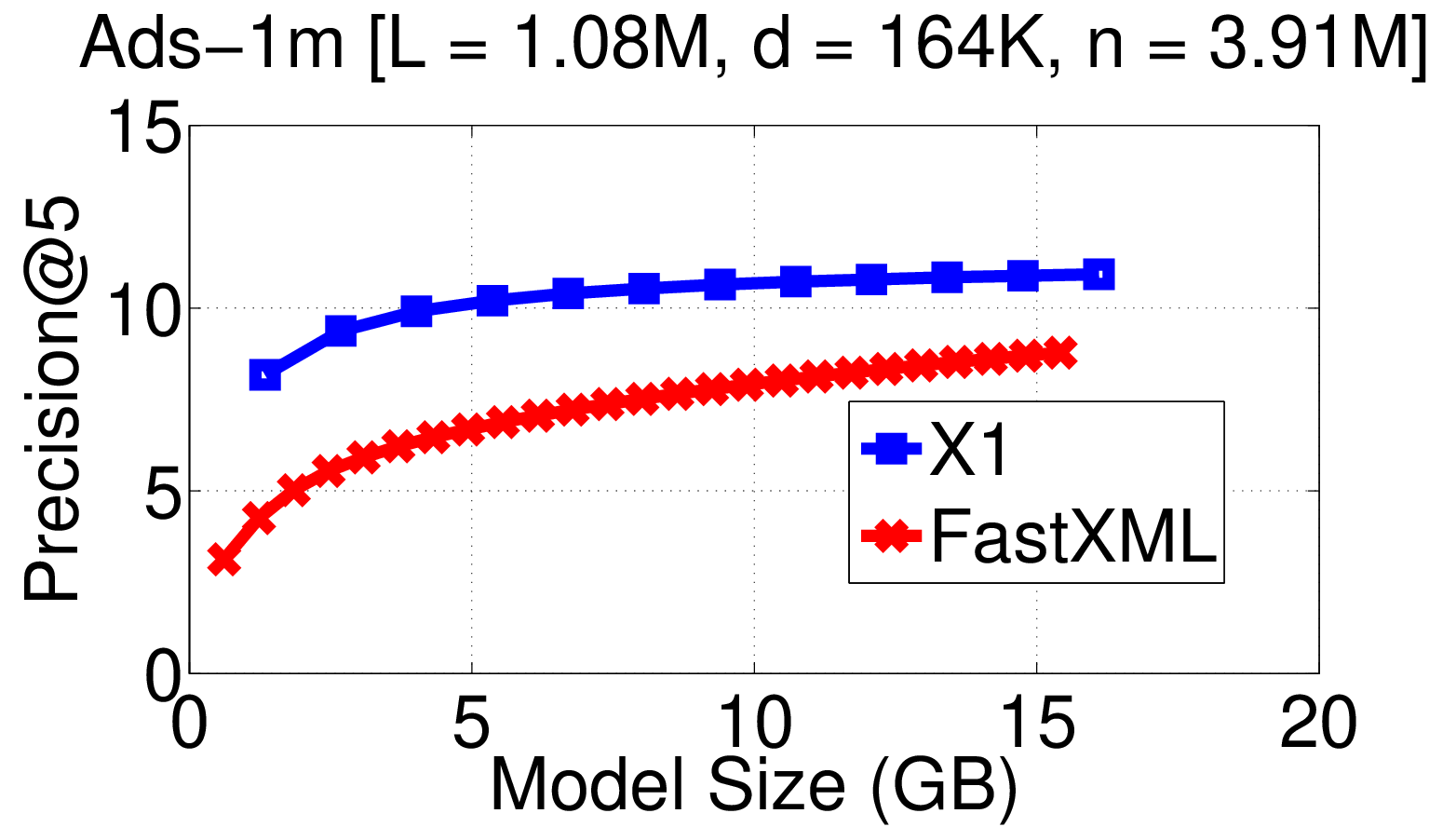}\\
(a)&(b)&(c)\\
\end{tabular}
\caption{{Variation of precision accuracy with model size on Ads-1m Data set}}
  \label{fig:ads-1mMP}
\end{figure}

\begin{figure}[h!]
\captionsetup{font=small} 
  \centering
\begin{tabular}{ccc}
  \includegraphics[width=.33\textwidth]{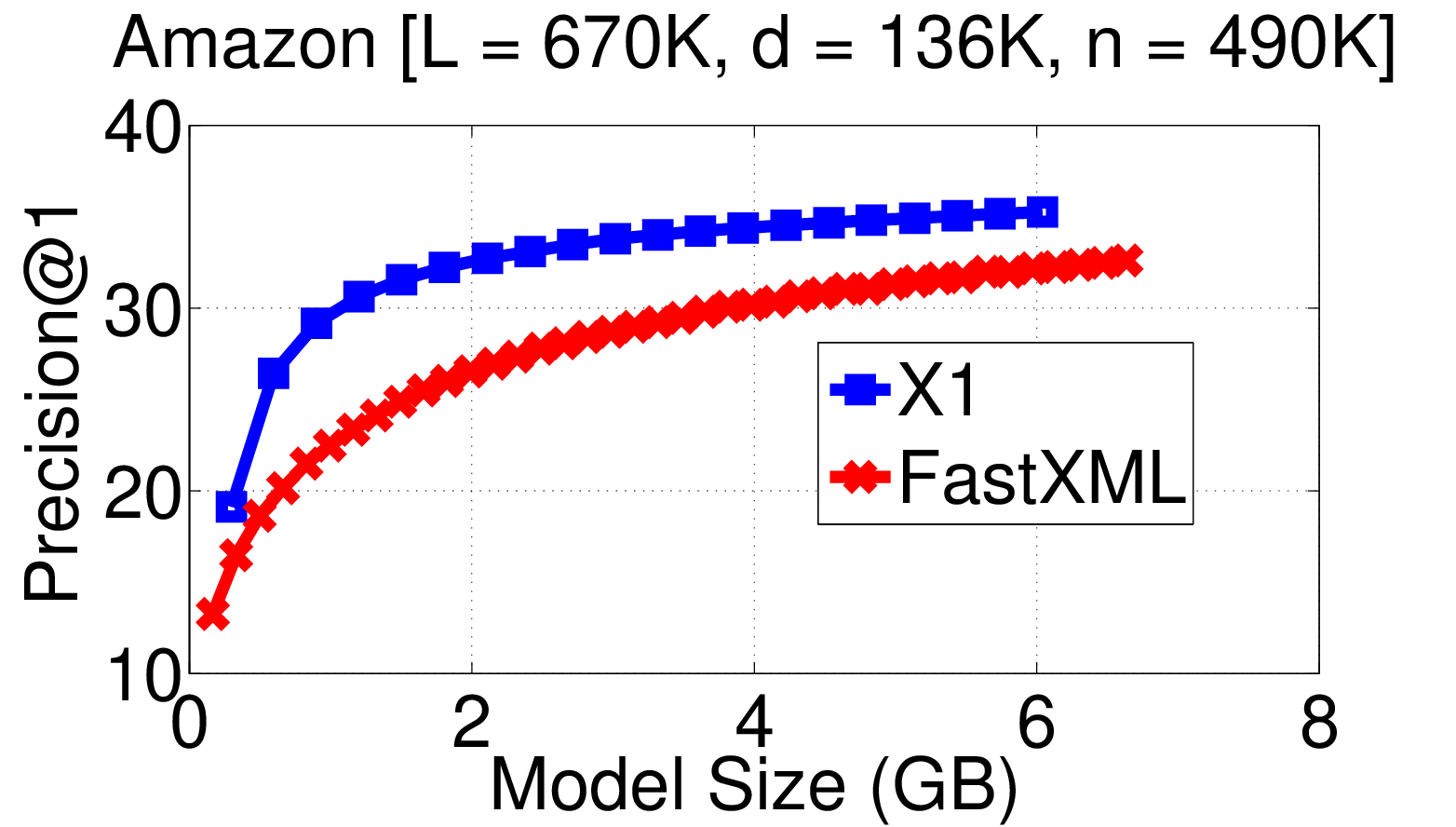}&
  \includegraphics[width=.33\textwidth]{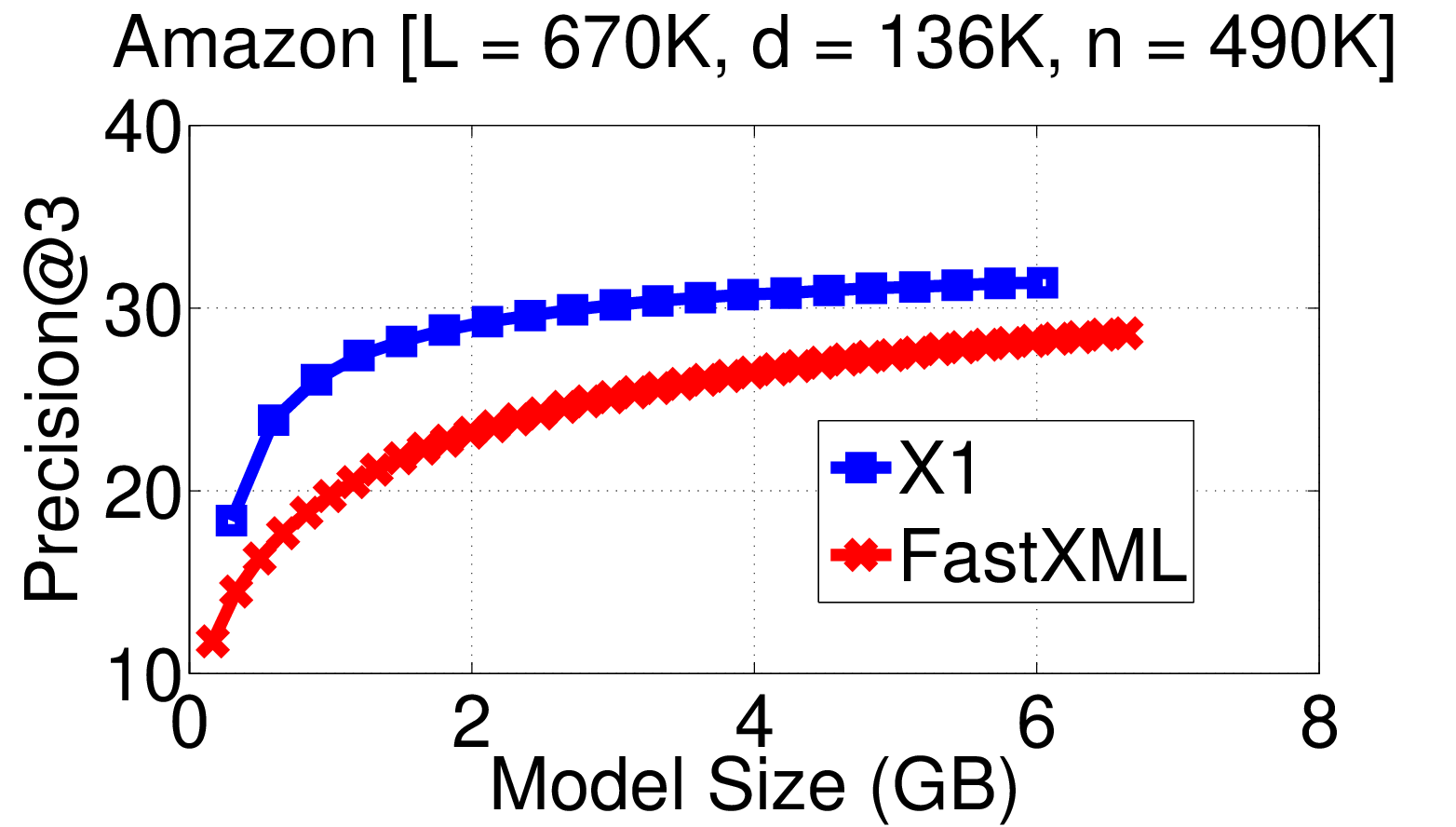}&
  \includegraphics[width=.33\textwidth]{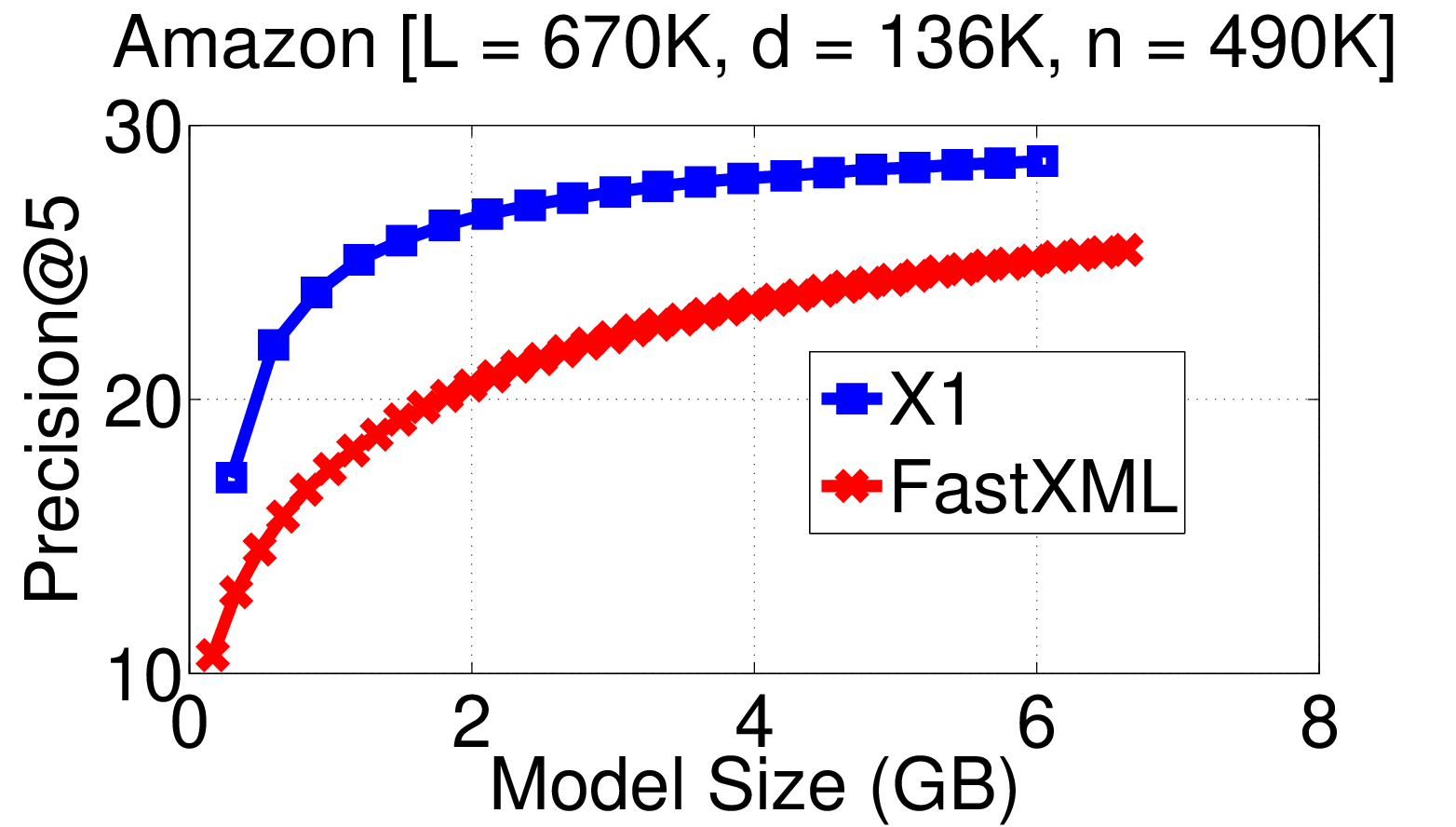}\\
(a)&(b)&(c)\\
\end{tabular}
\caption{\small{Variation of precision accuracy with model size on Amazon Data set}}
  \label{fig:amazonMP}
\end{figure}

\begin{figure}[h!]
\captionsetup{font=small} 
  \centering
\begin{tabular}{ccc}
  \includegraphics[width=.33\textwidth]{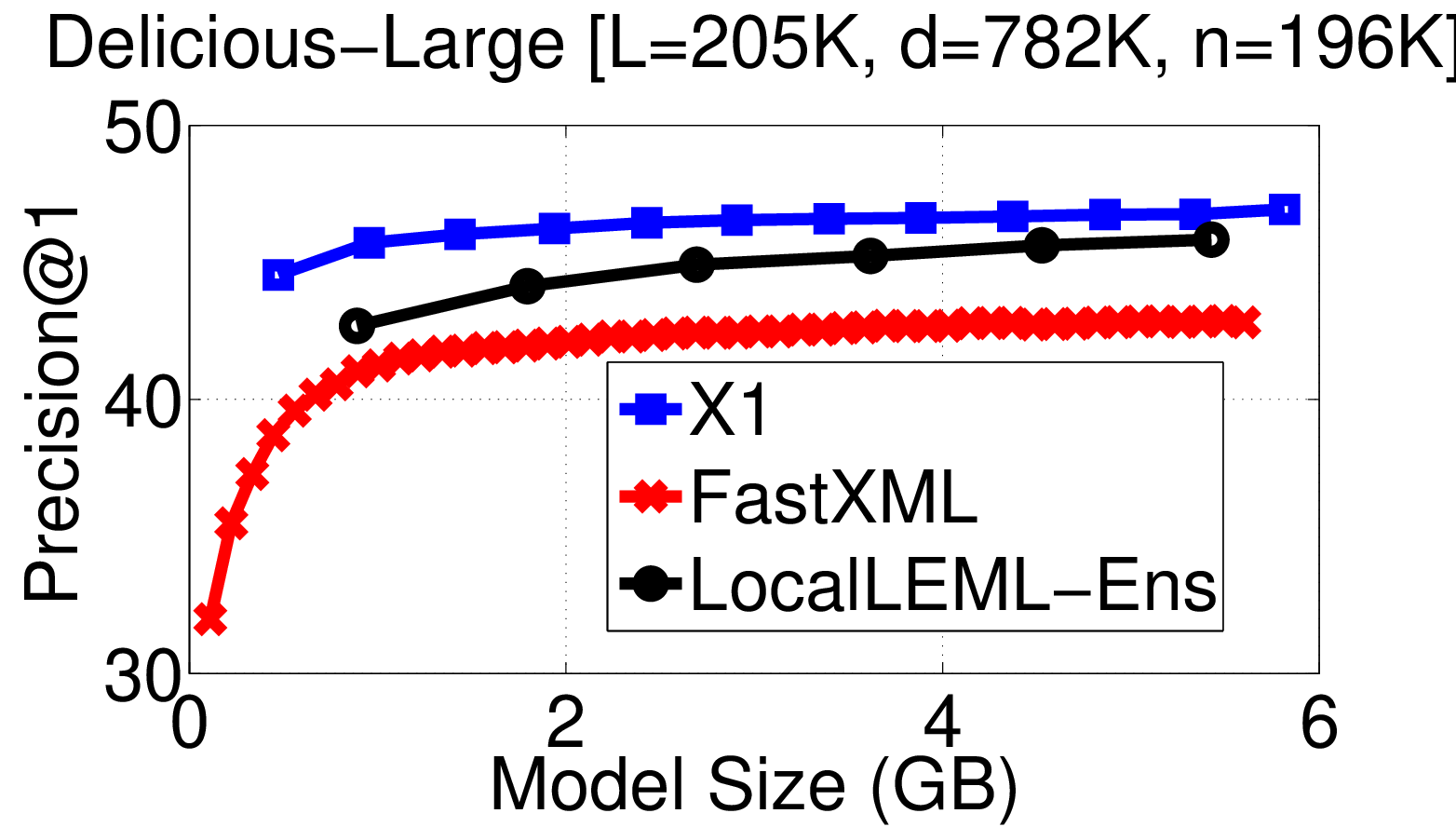}&
  \includegraphics[width=.33\textwidth]{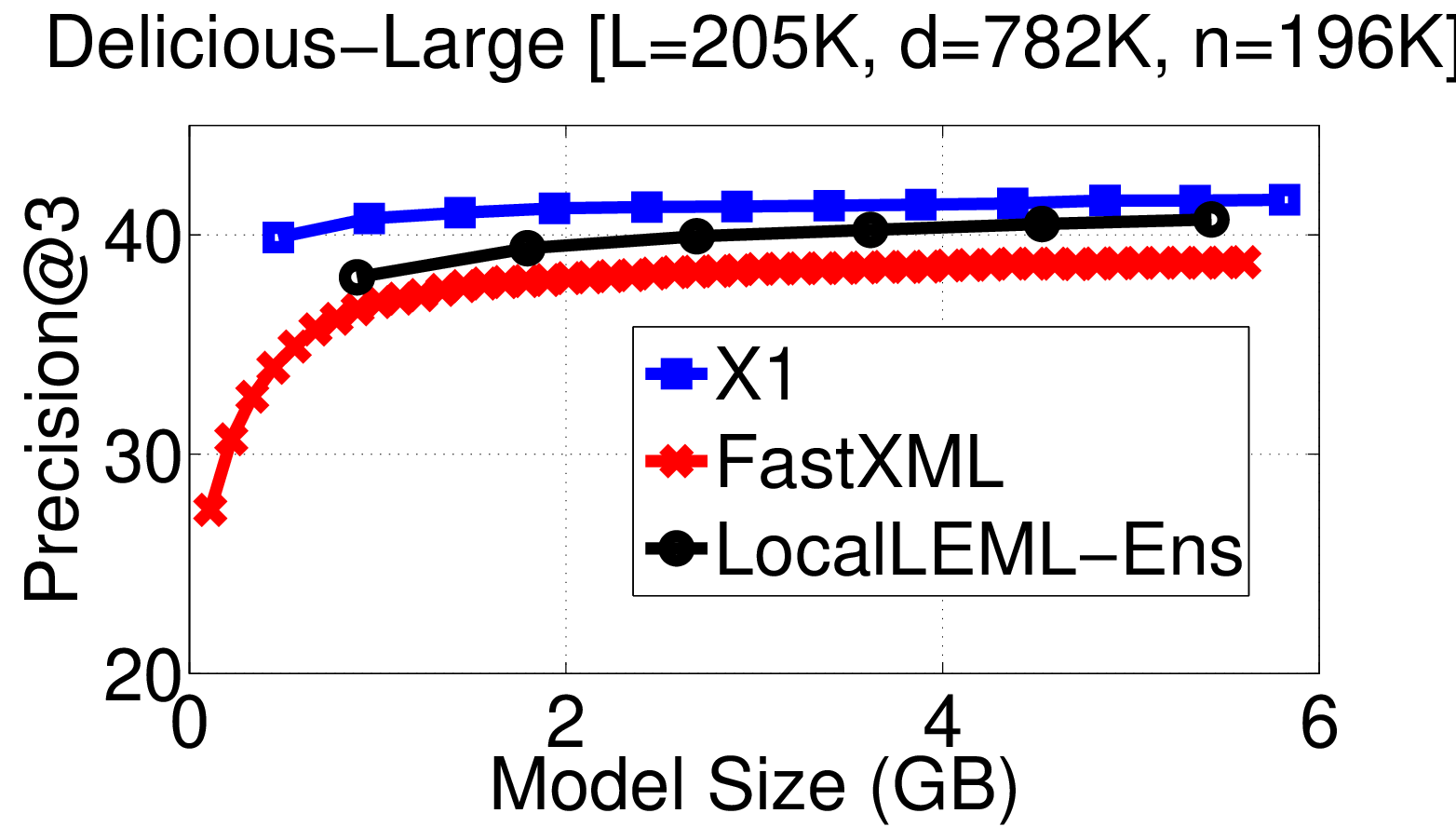}&
  \includegraphics[width=.33\textwidth]{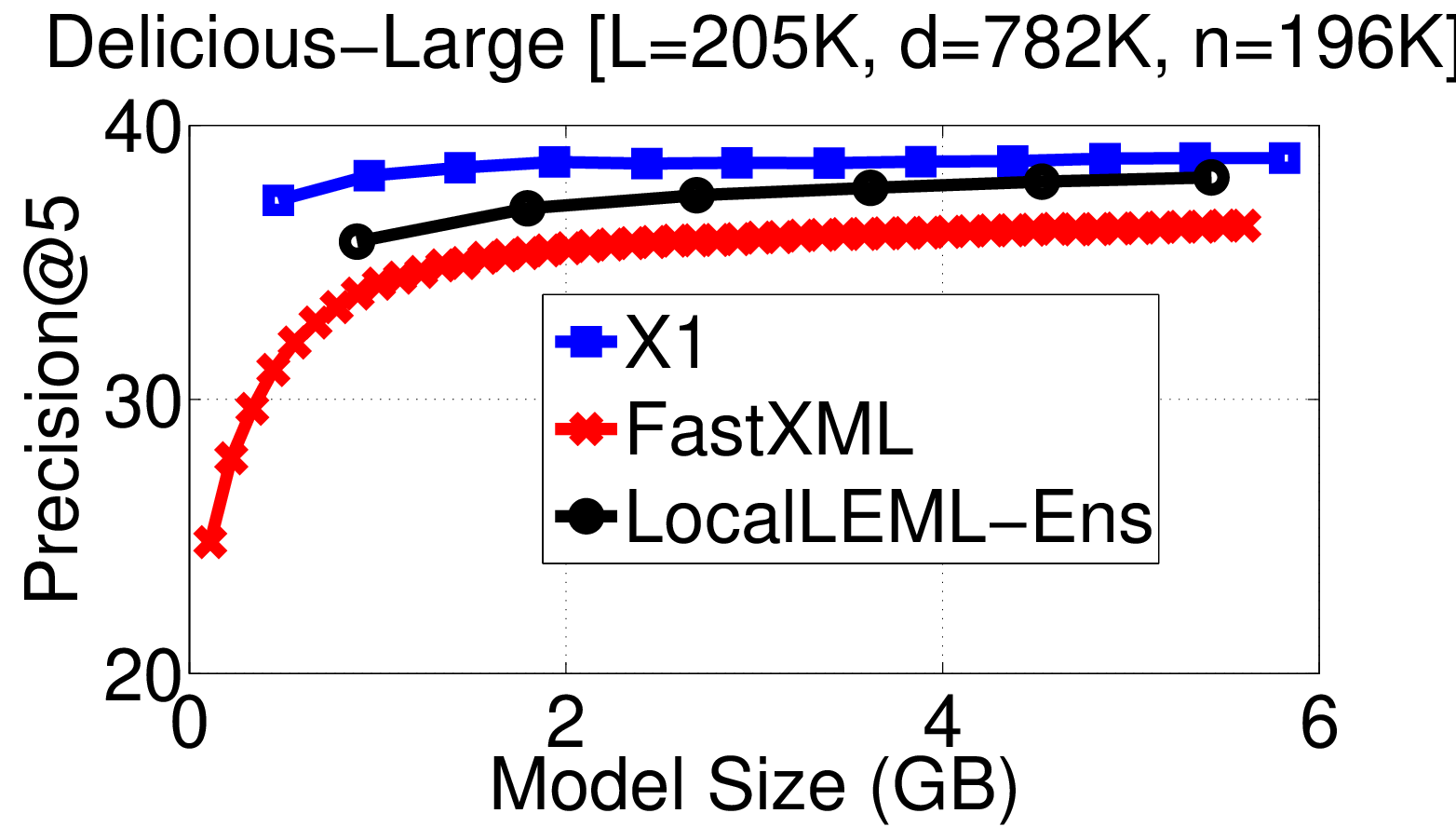}\\
(a)&(b)&(c)\\
\end{tabular}
\caption{\small{Variation of precision accuracy with model size on Delicious-Large Data set}}
  \label{fig:delLargeMP}
\end{figure}

\begin{figure}[h!]
\captionsetup{font=small} 
  \centering
\begin{tabular}{ccc}
  \includegraphics[width=.33\textwidth]{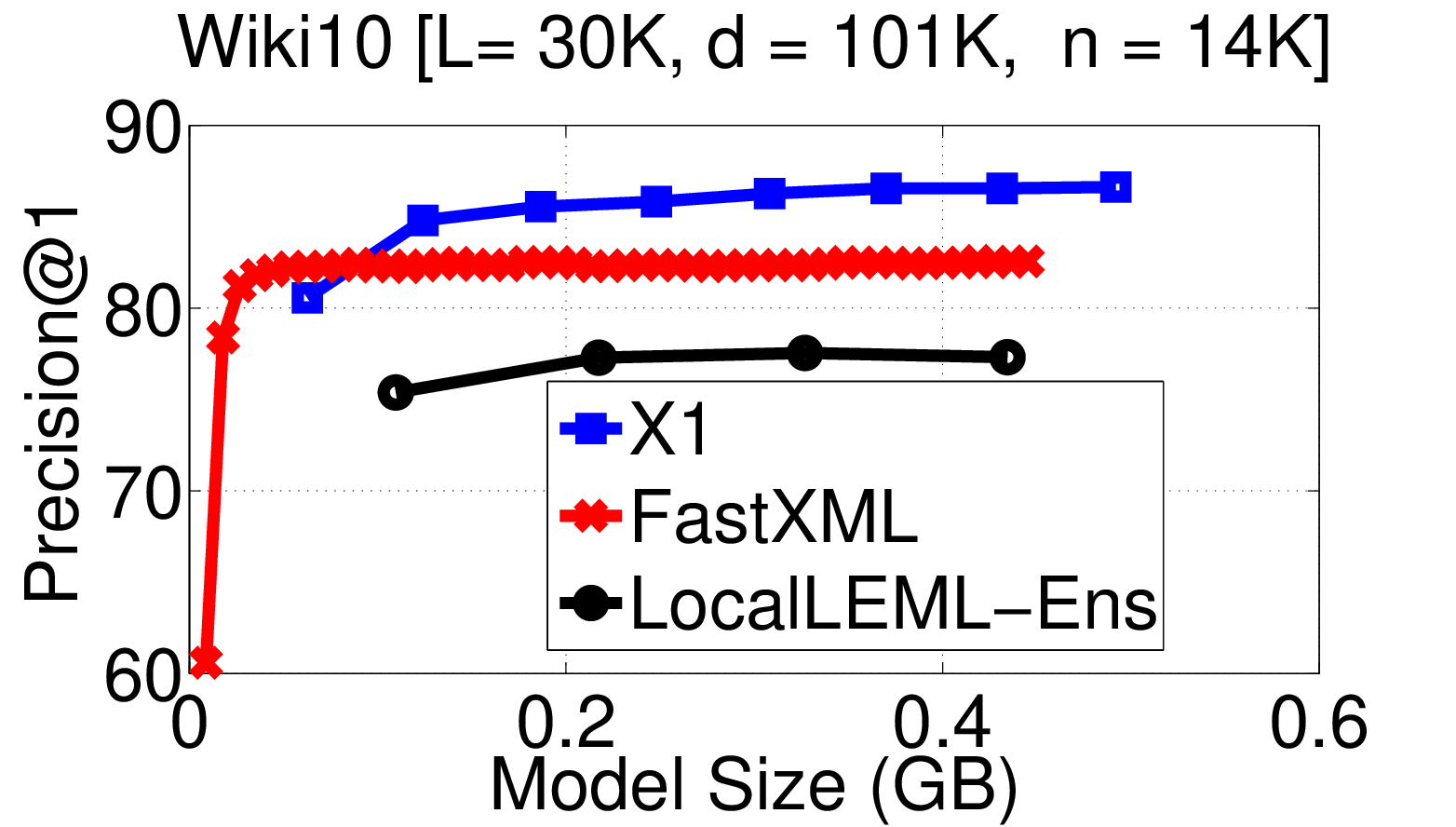}&
  \includegraphics[width=.33\textwidth]{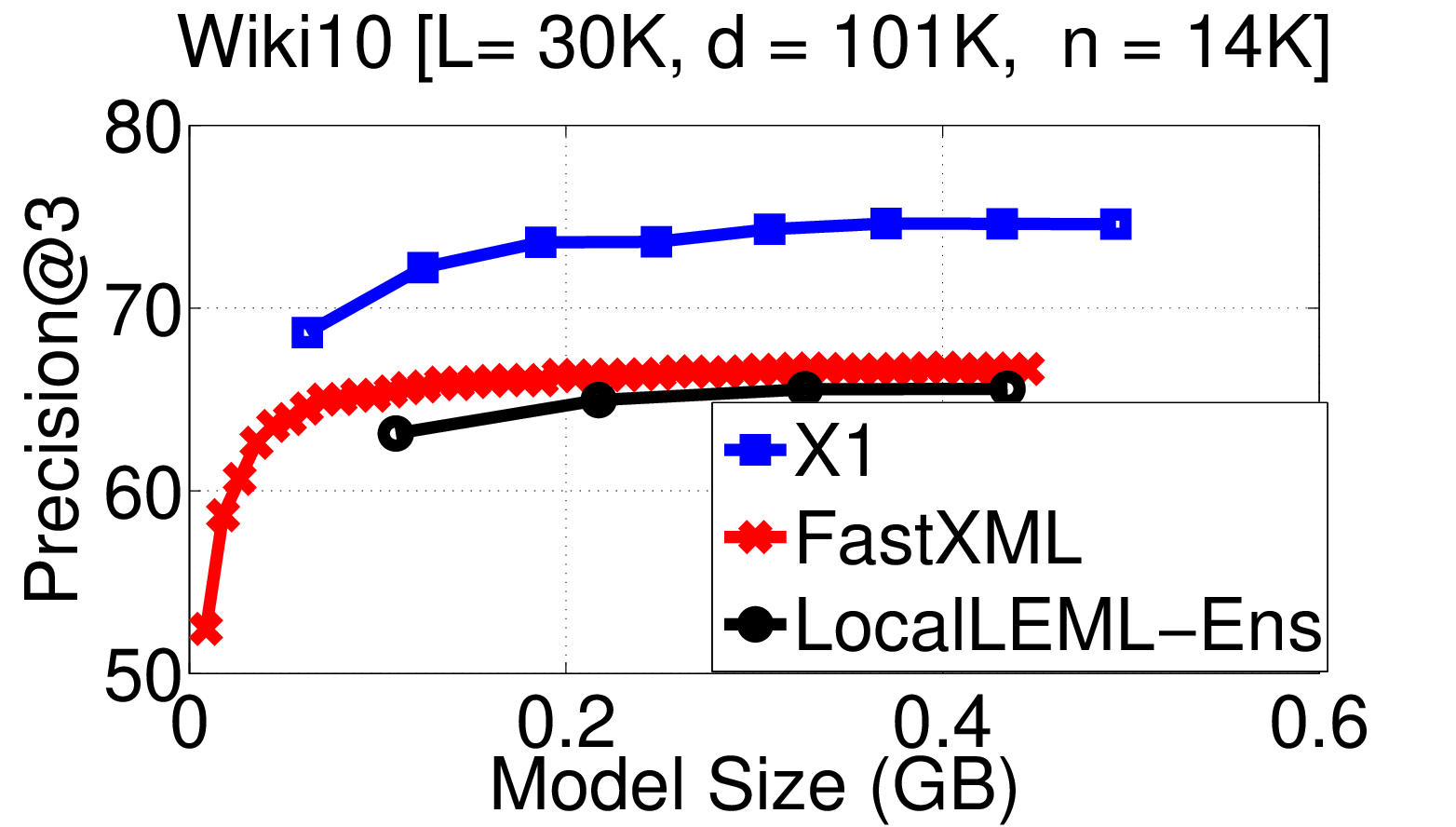}&
  \includegraphics[width=.33\textwidth]{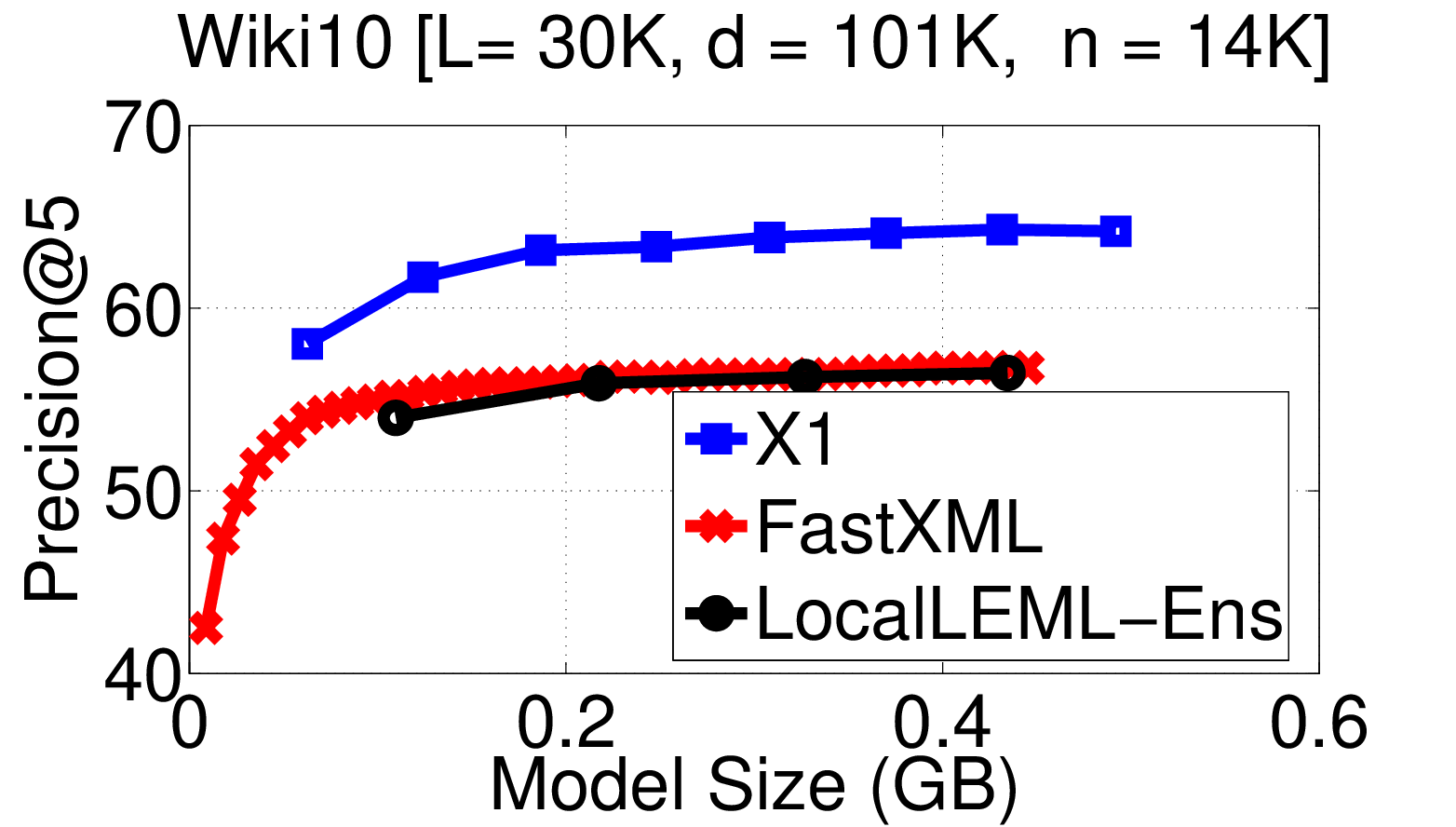}\\
(a)&(b)&(c)\\
\end{tabular}
 \caption{\small{Variation of precision accuracy with model size on Wiki10 Data set}}
  \label{fig:wiki10MP}
\end{figure}

\begin{figure}[h!]
\captionsetup{font=small} 
  \centering
\begin{tabular}{ccc}
  \includegraphics[width=.33\textwidth]{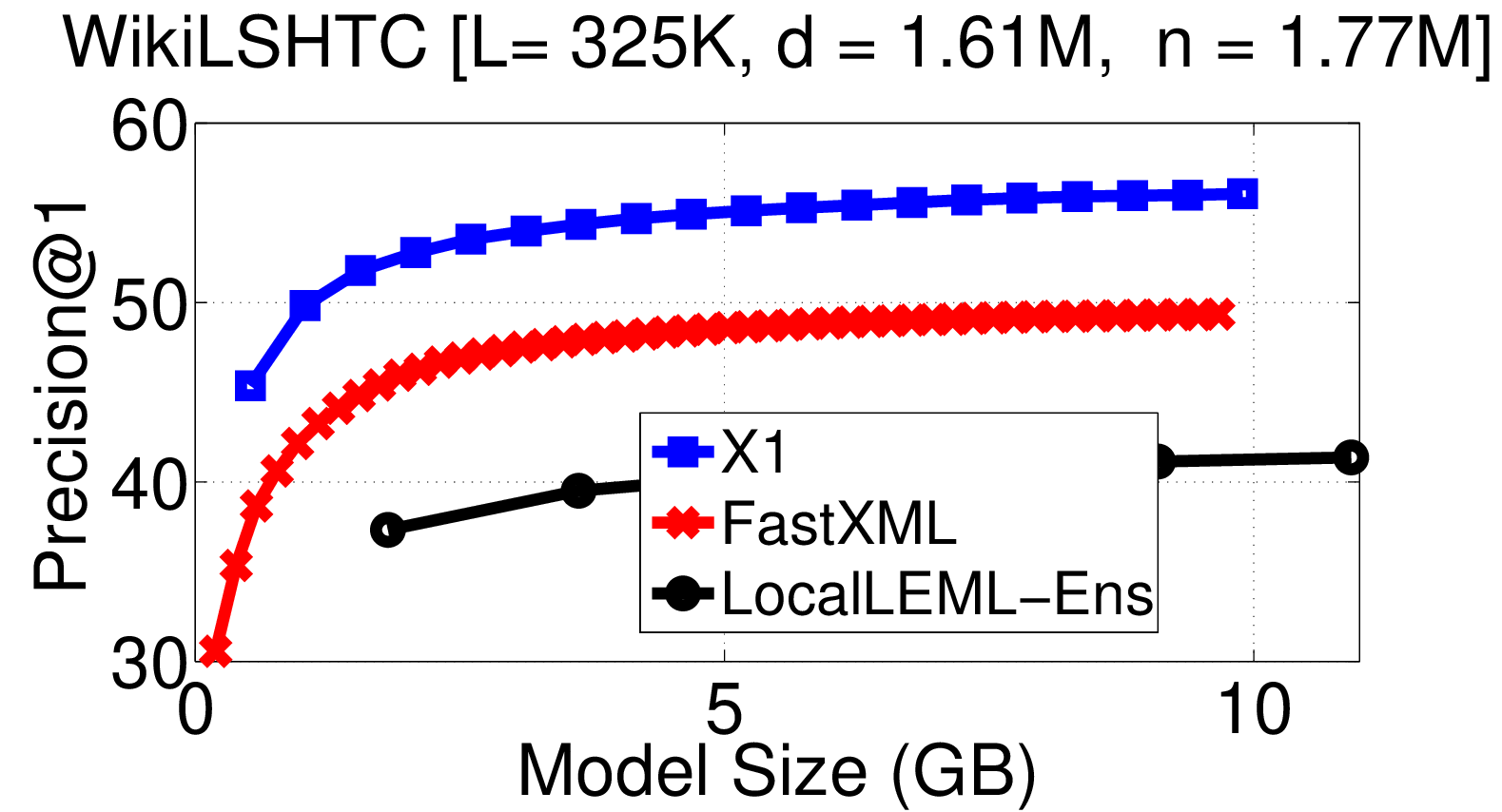}&
  \includegraphics[width=.33\textwidth]{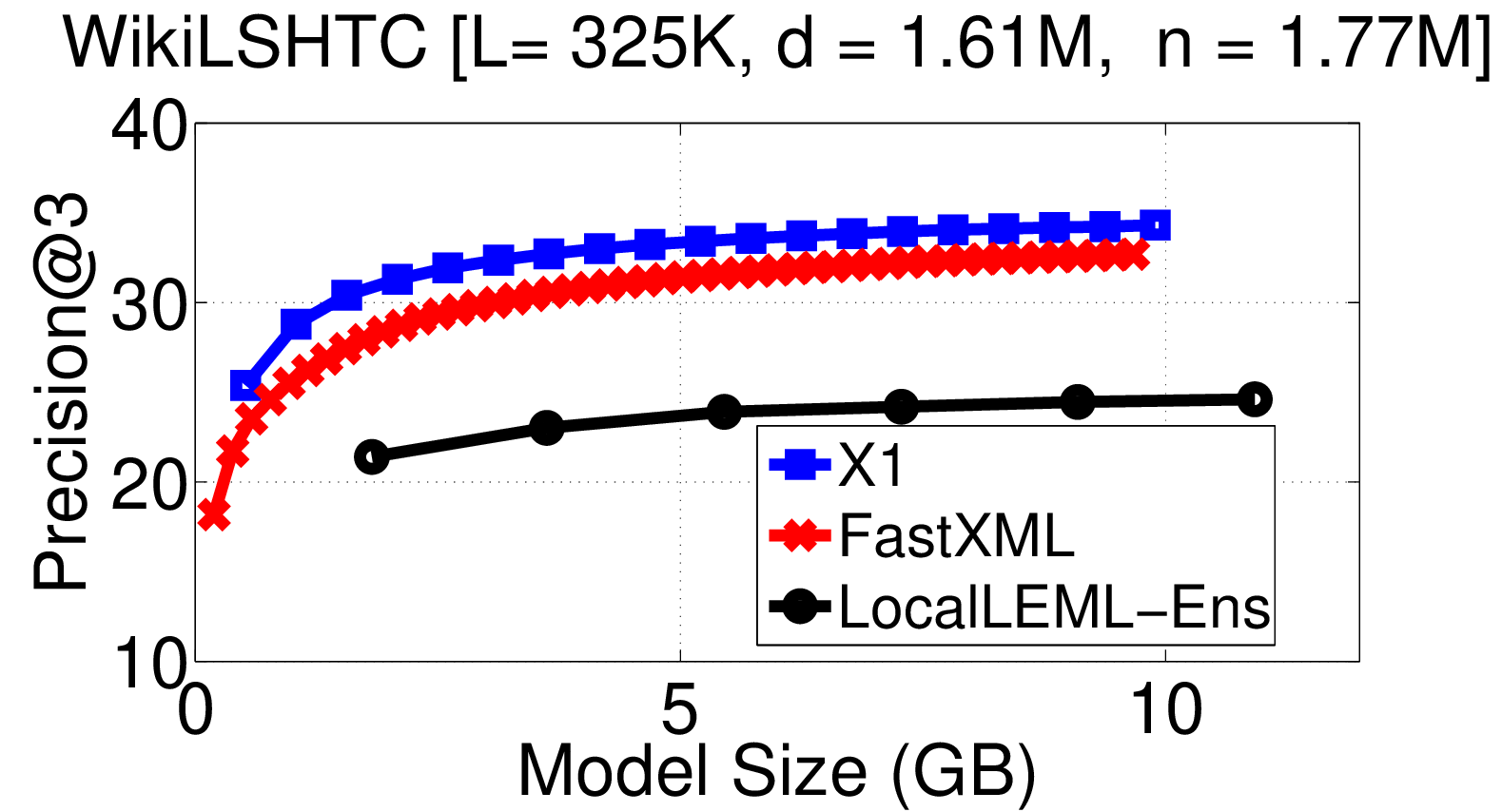}&
  \includegraphics[width=.33\textwidth]{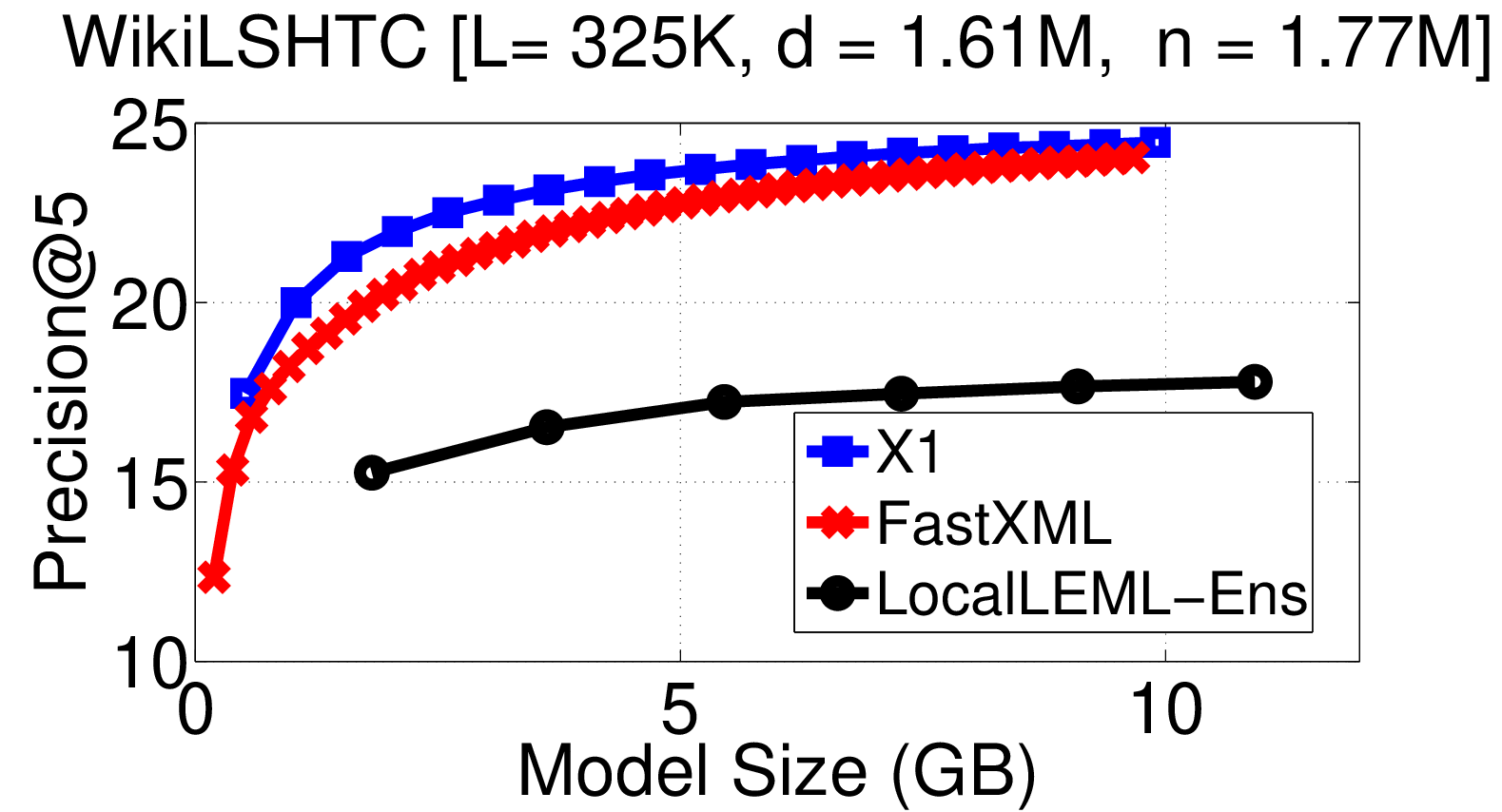}\\
(a)&(b)&(c)\\
\end{tabular}
\caption{\small{Variation of precision accuracy with model size on WikiLSHTC Data set}}
  \label{fig:wikiLSHTCMP}
\end{figure}

\end{document}